\documentclass[10pt,journal,compsoc]{IEEEtran}
\ifCLASSOPTIONcompsoc
  \usepackage{cite}
\else
  \usepackage{cite}
\fi
\ifCLASSINFOpdf
\else
\fi
\usepackage{amsmath,amsfonts}
\usepackage{amsthm,amssymb}
\usepackage{afterpage}
\usepackage{mathrsfs}
\usepackage{algorithmic}
\usepackage[hidelinks]{hyperref}
\usepackage{algorithm}
\usepackage{array}
\usepackage{hyperref}
\usepackage{float}
\ifCLASSOPTIONcompsoc
\usepackage[caption=false, font=normalsize, labelfont=sf, textfont=sf]{subfig}
\else
\usepackage[caption=false, font=footnotesize]{subfig}
\fi
\usepackage{cuted}
\usepackage{capt-of}
\usepackage{textcomp}
\usepackage{stfloats}
\usepackage{url}
\usepackage{verbatim}
\usepackage{graphicx}

\usepackage{cite}

\usepackage{soul}
\usepackage{bm}
\usepackage{booktabs}
\usepackage{txfonts}
\usepackage{booktabs}
\usepackage{multirow}
\usepackage{caption3}
\usepackage{caption}
\usepackage{xcolor}
\usepackage{float}
\usepackage[table]{xcolor}
\definecolor{darkgray}{rgb}{0.7, 0.7, 0.7}
\definecolor{gray}{rgb}{0.8, 0.8, 0.8}
\definecolor{lightgray}{rgb}{0.9, 0.9, 0.9}

\usepackage{pifont}
\usepackage{threeparttable}

\begin{document}
\title{LEVIRDet: A Million-Scale 159-Category Dataset and Foundation Model for Universal Remote Sensing Object Detection}

\author{Qinzhe~Yang, Dongyu~Wang, Haohan~Niu,~Jia~Xu,\\Zhenwei~Shi,~\IEEEmembership{Senior~Member,~IEEE},~and~Zhengxia~Zou$^\star$,~\IEEEmembership{Senior~Member,~IEEE}
\IEEEcompsocitemizethanks{
\IEEEcompsocthanksitem Qinzhe Yang and Haohan Niu are with Shen Yuan Honors College, Beihang University, Beijing 100191, China.
\IEEEcompsocthanksitem Dongyu Wang, Zhenwei Shi, and Zhengxia Zou are with the Department of Aerospace Intelligent Science and Technology, School of Astronautics, and with the State Key Laboratory of Virtual Reality Technology and Systems, Beihang University, Beijing 100191, China.
\IEEEcompsocthanksitem Jia Xu is with the Qian Xuesen
Laboratory of Space Technology, China Academy of Space Technology,
Beijing 100094, China. 
\IEEEcompsocthanksitem $^\star$ Corresponding author (zhengxiazou@buaa.edu.cn)
\IEEEcompsocthanksitem The work was supported by the National Natural Science Foundation of China under Grants 62125102, 62471014, U24B20177, U25A20401, and in part by the Fundamental Research Funds for the Central Universities.
}}

\markboth{Journal of \LaTeX\ Class Files,~Vol.~XX, No.~XX, XXX~XXXX}%
{Shell \MakeLowercase{\textit{et al.}}: Bare Demo of IEEEtran.cls for Computer Society Journals}

\IEEEtitleabstractindextext{%
\begin{abstract}

Remote sensing object detection has advanced rapidly with the development of large-scale benchmarks and modern detection architectures. However, existing datasets and detectors remain fragmented. Most benchmarks focus on limited categories, fixed spatial resolutions, or a single sensor, while detectors still struggle to work across different sensors and categorical systems. In this paper, we introduce LEVIRDet-159, the largest and most comprehensive remote sensing object detection dataset to date, with 159 categories, $\sim$2.56 million bounding boxes, and $\sim$700k fine-grained annotations under a multi-level taxonomy. In each key scale dimension, LEVIRDet-159 exceeds the corresponding largest existing remote sensing object detection dataset, containing approximately \(7\times\) more images, \(6\times\) more object instances, and \(4\times\) more categories (relative to the released FAIR1M with 393k instances and 37 categories). Based on this dataset, we design LEVIRDetNet, a scale-hierarchy-aware detection foundation model for universal remote sensing object detection. LEVIRDetNet couples online visual Ground Sampling Distance (GSD) prediction, GSD-conditioned query modulation and allocation, and a hierarchy-aware detection head for mixed-granularity remote sensing supervision. Under stringent evaluation settings, LEVIRDetNet demonstrates strong cross-domain generalization. Even without target-domain training or fine-tuning, it achieves state-of-the-art detection performance on 9 external benchmarks, improving the strongest fully supervised competing methods by 5.02 mean primary AP on average under each benchmark’s primary metric. It also remains strongest in score-threshold comparisons with open-set and grounding models, maintaining stable precision and recall at practical confidence thresholds. We hope this study will facilitate the development of strongly generalizable remote sensing object detection across diverse category systems, spatial resolutions, and sensor platforms. The full image tiles, annotations, source-license manifest, code, and trained models have been released in a versioned project repository at \url{https://qinzheyang.github.io/LEVIRDet/}.
\end{abstract}
}

\maketitle

\IEEEdisplaynontitleabstractindextext
\IEEEpeerreviewmaketitle

\section{Introduction}\label{sec:introduction}

\begin{figure*}[!t]
\centering
\includegraphics[width=1\linewidth]{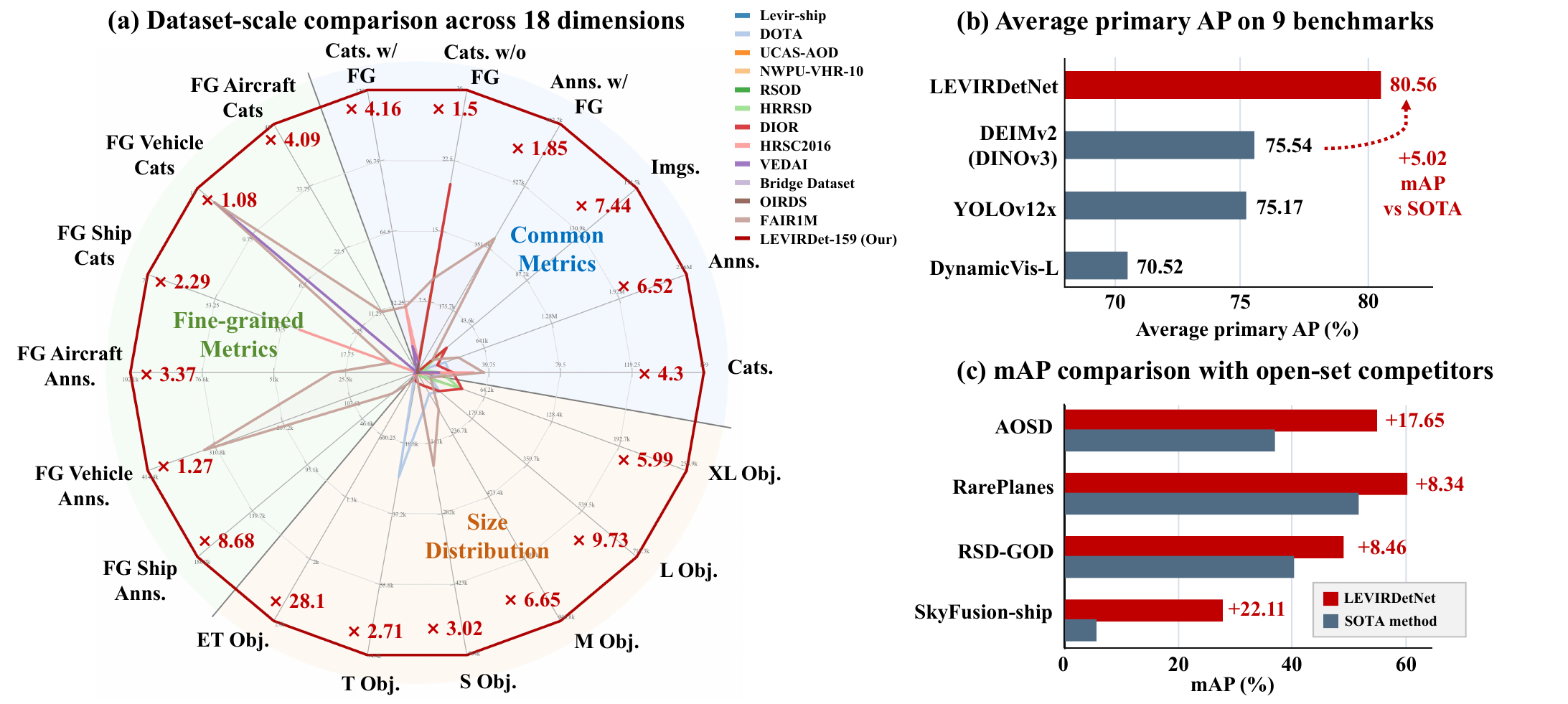}
    \vspace{-4ex}
\caption{
Overview of target-training-free cross-benchmark performance and dataset-scale comparison. (a) Dataset-scale comparison for LEVIRDet-159 dataset. (b) Average primary AP on 9 benchmarks. (c) mAP comparison with open-set competitors. The figure indicates the scale of each dataset, and LEVIRDet-159 has reached the largest scale in 18 dimensions. FG, Anns., Cats., and Obj. denote fine-grained, annotations, categories, and objects, respectively. ET, T, S, M, L, and XL denote extremely tiny, tiny, small, medium, large, and extra-large objects, respectively. Object size is measured by the square root of bounding-box area: ET $<$ 4 pixels, T $\in [4,8)$ pixels, S $\in [8,16)$ pixels, M $\in [16,32)$ pixels, L $\in [32,96)$ pixels, and XL $\geq$ 96 pixels.}
\label{fig:ModelOverview}
\vspace{-2ex}
\end{figure*}

\IEEEPARstart{R}{emote} sensing object detection is a fundamental task for Earth observation, supporting applications such as urban monitoring, transportation analysis, and maritime surveillance. It has advanced rapidly with the development of large-scale benchmarks and modern detection architectures~\cite{ObjectDetectioninAerialImagesTPAMI,Objectdetectionin20years}. Existing benchmarks have expanded category coverage and annotation density, while transformer-based detectors and visual foundation models have improved end-to-end detection and cross-dataset transfer~\cite{dota,DIOR,fair1m}. However, these advances have not yet led to a universal detector that can reliably operate across sensors, resolutions, and category systems.

This gap comes from both data and model limitations. On the data side, existing benchmarks usually cover only a narrow portion of the remote sensing detection space~\cite{SmallObjectTPAMI,levirship,dota,ucasaod,NWPU,rsod,hrssd,tas,DIOR,HRSC2016,vedai,ITCVD,DLR3kDLR,bridge,OIRDS,mar20,fair1m,vhrships,shiprsimagenet,Ningbo,hrplane}. Some datasets focus on a small set of common categories, some are designed for a single object family such as aircraft, ships, or vehicles~\cite{ucasaod,rsod,hrssd,tas,bridge,OIRDS,Ningbo,hrplane}, and others emphasize specific settings such as tiny-object detection or fine-grained recognition~\cite{SmallObjectTPAMI,levirship,dota,HRSC2016,vedai,ITCVD,DLR3kDLR,mar20,fair1m,vhrships,shiprsimagenet}. Their image collections are often limited in scale, tied to fixed spatial resolutions or a narrow GSD range, and dominated by one sensor, platform, or imaging style. As a result, detectors trained on these benchmarks are exposed to limited category vocabularies, object-size distributions, scene densities, and sensor appearances, which weakens their ability to transfer across real-world remote sensing scenarios. In addition, their category definitions, bounding-box protocols, and image resolutions are often inconsistent. Directly aggregating them can introduce conflicting supervision, where visually similar objects are annotated with different box conventions, category granularities, or instance completeness across sources. 

On the model side, most remote sensing detectors still inherit general natural-image detection designs, including two-stage, one-stage, and DETR-like architectures. These models rarely encode domain-specific factors such as ground sampling distance (GSD), image-dependent scene density, or ancestor-descendant relations among mixed-granularity categories, making them difficult to transfer across different sensors, spatial resolutions, and categorical systems. We argue that a practical remote sensing detection foundation model requires not only larger training data, but also model structures that reflect the scale, density, and semantic hierarchy of remote sensing imagery.

To address these issues, we present LEVIRDet\footnote{LEVIR refers to the authors' laboratory, namely the LEarning, VIsion, and Remote Sensing Laboratory.}, a universal remote sensing object detection framework consisting of LEVIRDet-159 and LEVIRDetNet. LEVIRDet-159 is a 159-category, million-scale dataset with $\sim$\(1.8\times10^5\) images, $\sim$2.56 million bounding boxes, and $\sim$700k fine-grained annotations under a multi-level taxonomy. Compared with representative public remote sensing detection datasets\cite{levirship,dota,ucasaod,NWPU,rsod,hrssd,tas,DIOR,HRSC2016,vedai,ITCVD,DLR3kDLR,bridge,OIRDS,mar20,fair1m,vhrships,shiprsimagenet,Ningbo,hrplane}, it is substantially larger in image count (\(\times7\)), annotation count (\(\times6\)), and category coverage (\(\times4\)), making it, to our knowledge, the first million-box dataset built for broad-category universal remote sensing object detection under a unified tight horizontal bounding box (tight-HBB) protocol. Beyond scale, LEVIRDet-159 is organized around semantic observability: each instance is labeled at the most specific category supported by visual evidence, while ambiguous objects can fall back to reliable ancestor categories.

LEVIRDet-159 is derived from heterogeneous public remote sensing resources\cite{levirship,dota,ucasaod,NWPU,rsod,hrssd,tas,DIOR,HRSC2016,vedai,ITCVD,DLR3kDLR,bridge,OIRDS,mar20,fair1m,vhrships,shiprsimagenet,Ningbo,hrplane}, but it is not a direct merge of their original labels. Instead, we rebuild the annotations through a systematic data engine that standardizes sources with tight horizontal boxes, a multi-level taxonomy, and source-controlled split rules. A large portion of the final annotations are newly produced or revised, including $\sim$1.48 million newly added boxes and $\sim$0.85 million boxes refined through geometry correction or fine-grained relabeling. This process converts heterogeneous resources into consistent, dense, and semantically reliable supervision, while preserving the limits of visual evidence under varying GSD and imaging conditions. By a conservative estimate, it required over 10k person-hours of human annotation and quality control.

Based on LEVIRDet-159, we design LEVIRDetNet, a scale-hierarchy-aware detection foundation model for universal remote sensing object detection. LEVIRDetNet adapts an end-to-end detector to remote sensing by coupling online visual GSD conditioning, dynamic query allocation, and a hierarchy-aware detection head. The GSD branch provides an image-level physical scale cue for query modulation, the dynamic query module adjusts object hypotheses to scene density, and the hierarchy-aware head learns from mixed-granularity labels while remaining compatible with flat benchmark evaluation. These designs jointly address the large resolution span, heterogeneous imaging appearance, dense small-object scenes, and complex category structure of remote sensing imagery. Following the task-focused precedent of SAM~\cite{sam3}, we refer to LEVIRDetNet as a detection foundation model because it is trained on large-scale, heterogeneous imagery and generalizes across remote sensing scenes and data sources without benchmark-specific fine-tuning. 

We evaluate LEVIRDetNet in a target-training-free cross-benchmark setting. LEVIRDetNet is trained once on LEVIRDet-159 and then evaluated on the test sets of 9 external benchmarks without benchmark-specific fine-tuning. Three benchmarks are source-seen but test-disjoint, while the remaining six are source-disjoint with respect to LEVIRDet-159 detector training. In contrast, the closed-set competing detectors in Fig.~\ref{fig:ModelOverview}(b) follow the standard target-supervised protocol and are trained or fine-tuned on the corresponding target benchmarks. Under this stricter setting, LEVIRDetNet achieves an mean primary AP of 80.56 across benchmarks covering aircraft, vehicles, ships, and general aerial objects, compared with 75.54 for the strongest competing method, outperforming 5.02 mean primary AP. When compared with the strongest target-supervised competitor on each benchmark and averaged over benchmarks, LEVIRDetNet still brings a 3.33 mean primary AP gain. Threshold-sweep comparisons further show stronger precision-recall stability than open-set and grounding models under practical confidence thresholds. 

The contributions of this paper can be summarized as follows:

1) We construct LEVIRDet-159, the up-to-date largest and most fine-grained remote sensing object detection dataset, containing 174,488 images, over 173.5 billion pixels, and 2,563,973 instances. It covers 30 common parent categories and 159 category types across global regions and diverse imaging conditions, and surpasses existing datasets in category number (\(\times4\)), image number (\(\times7\)), annotation number (\(\times6\)), object-size coverage (\(\times 2.71\) \(\sim\) \(\times 28.1\)), and geographic coverage.

2) We propose LEVIRDetNet, a scale-hierarchy-aware detection foundation model designed for universal remote sensing detection. Built upon a unified data engine with tight horizontal boxes, multi-level taxonomy, and visual-evidence-aware mixed-granularity labels, LEVIRDetNet explicitly models the coupling among object scale, scene density, and semantic hierarchy. It introduces online visual GSD conditioning for scale-aware query modulation, dynamic query allocation for image-adaptive object hypotheses, and hierarchy-aware classification for learning from mixed-granularity supervision while remaining compatible with flat benchmark evaluation.

3) Comprehensive experiments conducted on nine external benchmarks, including CarPK\cite{carpk}, NWPU\cite{NWPU}, VHRV\cite{vhrv}, SkyFusion\cite{skyfusion}, among others\cite{corsadd,hrplane,adcos,ucasaod}, reveal that LEVIRDetNet achieves state-of-the-art target-training-free cross-benchmark performance compared with existing closed-set, open-set, and grounding-based detectors. Without target-domain training or fine-tuning, LEVIRDetNet ranks first on all nine benchmarks and improves the strongest competing methods by 5.02 mean primary AP on average, while maintaining stronger precision-recall stability under practical confidence thresholds. These results demonstrate not only higher detection accuracy, but also strong generalization across diverse category systems, spatial resolutions, and sensor platforms, which is a key property of a remote sensing detection foundation model.

\section{Related Work}

\subsection{Object Detection}

Existing methods can be broadly classified into traditional dense-prediction detectors and query-based detectors\cite{rs4d}. Representative dense-prediction detectors include Faster R-CNN \cite{FasterR-CNN}, Cascade R-CNN \cite{Cascader-cnn}, YOLO series \cite{yolo}, SSD \cite{ssd}, RetinaNet \cite{retinanet}, FCOS \cite{fcos}, and CenterNet \cite{centernet}, while query-based detectors include DETR \cite{detr}, Deformable DETR \cite{deformable_detr}, DINO \cite{dino}, RT-DETR \cite{rtdetr}, and DEIM-style detectors \cite{deim}.

Recent advances in visual foundation models and open-vocabulary detection have catalyzed significant progress. Methods such as Grounding DINO\cite{groundingdino}, LAE-DINO\cite{laedino}, and OWL-ViT\cite{OWLVIT} introduce language prompts or vision-language pre-training to improve category transfer. However, these models are still sensitive to small objects, prompt wording, score thresholds, and duplicate predictions in remote sensing scenes. Moreover, they do not directly resolve inconsistent annotation protocols, scale-dependent visual observability, or mixed-granularity labels. In contrast, our work focuses on universal remote sensing detection by jointly constructing a unified large-scale dataset and designing a detector that explicitly models scale, density, and semantic hierarchy.

\subsection{Remote Sensing Detection Datasets}

Remote sensing object detection has been greatly advanced by large-scale benchmark datasets\cite{ObjectDetectioninAerialImagesTPAMI,GLHTPAMI}. Early datasets such as NWPU VHR-10\cite{NWPU} and HRSC2016\cite{HRSC2016} provide standard evaluation settings for general aerial objects and ships. DOTA\cite{dota} promotes oriented object detection in large aerial images, while DIOR\cite{DIOR} expands category diversity for horizontal detection, and FAIR1M\cite{fair1m} focuses on fine-grained oriented object detection. Other resources, including ShipRSImageNet\cite{shiprsimagenet}, RarePlanes\cite{rareplanes}, and CarPK\cite{carpk}, study family-specific recognition, tiny objects, or vehicle counting. LAE-1M~\cite{laedino} combines existing annotations and automatically labeled imagery for open-vocabulary detection, with roughly 1,600 vocabulary entries and one million labeled objects reported in its paper.

Despite their progress, existing remote sensing detection datasets remain insufficient for training a universal detection foundation model. Many benchmarks are designed for specific object families or task settings, while general-purpose datasets still have limited category coverage and semantic granularity \cite{levirship,ucasaod,MTARSI,rsod,hrssd,HRSC2016}. Their resolution range and source diversity are also constrained, making it difficult to capture the scale variation and imaging diversity of real remote sensing applications. As a result, detectors trained on these datasets often learn dataset-specific priors rather than broadly transferable object semantics. To address these limitations, we construct LEVIRDet-159 as a large-scale universal detection dataset with broad categories, multi-source imagery, and a multi-level taxonomy, providing a stronger data foundation for generalizable remote sensing object detection.


\subsection{Scale Variation and Fine-Grained Detection}

Scale variation, object density, and category hierarchy are central challenges in remote sensing detection\cite{fair1m,rs4d}. Existing methods usually handle scale variation through feature or image pyramids, multi-scale training, and tiny-object enhancement strategies\cite{fpn}. Dense scenes are often addressed by improving assignment strategies, proposal generation, or post-processing\cite{NWD,SmallObjectTPAMI,SMALLTPAMI}. Fine-grained and hierarchical recognition methods further exploit category structures to distinguish visually similar objects and organize predictions at different semantic levels\cite{HRSC2016,shiprsimagenet}.

However, most existing methods treat scale, density, and hierarchy as separate problems. Scale is mainly modeled at the feature level, density is handled by proposal or query design, and hierarchy is often used only for label organization or post-hoc grouping\cite{rssurvey}. This decoupled treatment is insufficient for universal remote sensing detection, where the reliable semantic depth of an object depends on GSD, imaging quality, object size, and scene context. Therefore, we propose LEVIRDetNet, which addresses these factors through dedicated but coordinated components, including online visual GSD estimation, dynamic query allocation, and hierarchy-aware classification to jointly model scale, density, and semantic granularity.

\section{Dataset and Data Engine}
\label{sec:data_engine_dataset}

\subsection{LEVIRDet-159 Dataset}
\label{sec:levir_det_159}

LEVIRDet-159 is built for universal remote sensing object detection rather than for a single object family and sensor. Its central design is to couple cross-source image diversity, tight dense localization, and semantic labels whose depth depends on visual evidence. This coupling is important because a universal detector must decide not only where objects are, but also how many object hypotheses are needed in a scene and how fine a category prediction can be trusted under the available GSD and image quality.

\begin{figure*}
\centering
\includegraphics[width=1\linewidth]{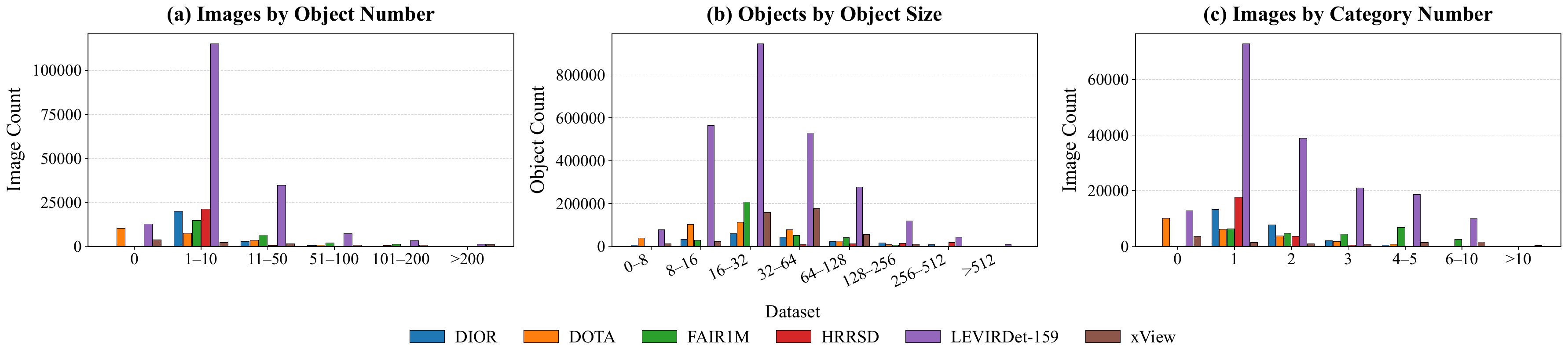}
\vspace{-4ex}
\caption{
Comparison of dataset statistics. The three panels compare image counts by object number, object counts by object size, and image counts by category number. Comparison datasets include DIOR \cite{DIOR}, DOTA \cite{dota}, FAIR1M\cite{fair1m}, HRRSD\cite{hrssd}, and xView\cite{xview}.
}
\label{fig:combined_dataset_statistics}
\vspace{-2ex}
\end{figure*}

\subsubsection{Overall Scale}


\begin{figure}
\centering
\includegraphics[width=0.85\linewidth]{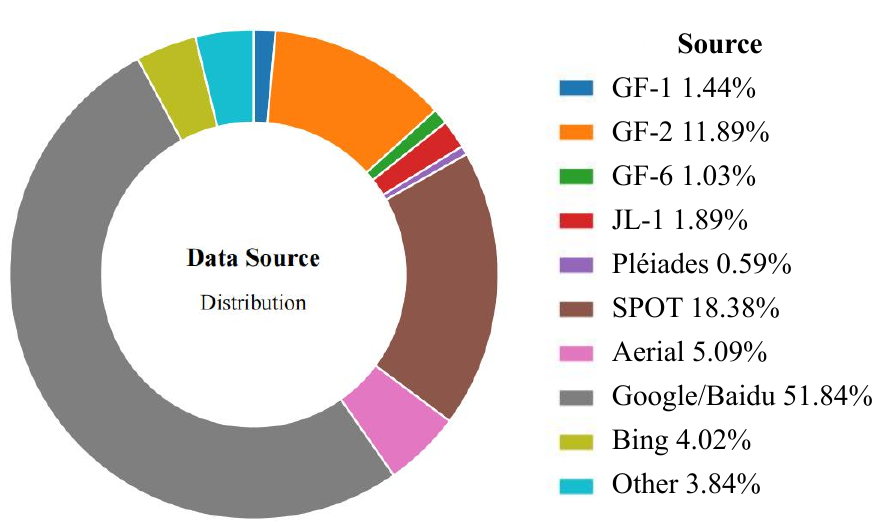}
\vspace{-2ex}
\caption{
An image source distribution of LEVIRDet-159, covering more than ten satellite, aerial, and map-service sources.
}
\label{fig:data_source_donut_times_larger}
\vspace{-2ex}
\end{figure}

\begin{figure*}
\centering
\includegraphics[width=1\linewidth]{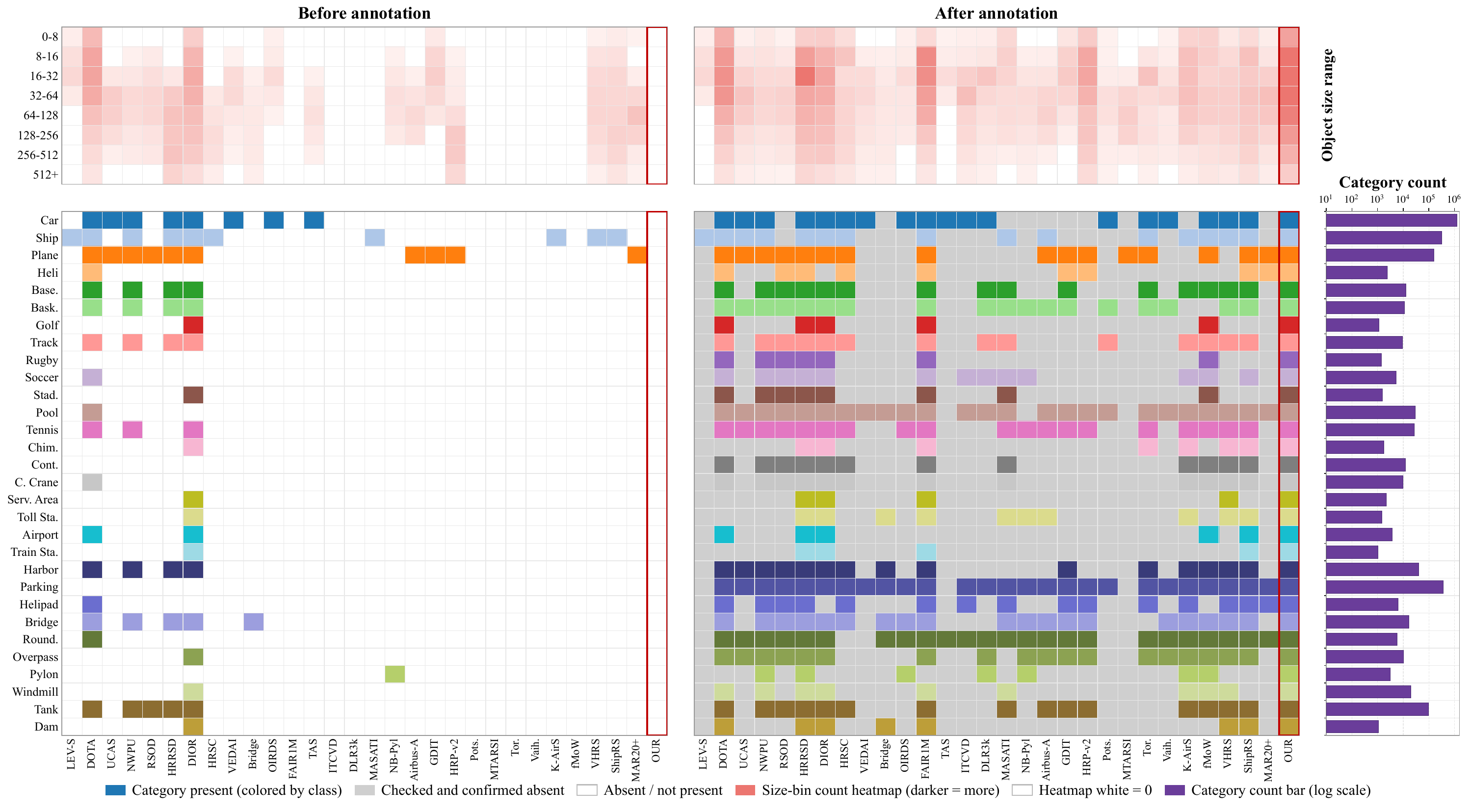}
\vspace{-4ex}
\caption{
Category and annotation expansion of LEVIRDet-159. The figure compares category coverage before and after unification, summarizes annotation changes, category counts, and the object-size distribution heatmap.}
\label{fig:dataset_category_before_after_with_size_and_curve}
\vspace{-2ex}
\end{figure*}

To the best of our knowledge, LEVIRDet-159 is up-to-date the largest and most fine-grained remote sensing object detection dataset, containing 174,488 images, over 173.5 billion pixels, and 2,563,973 instances. It covers 30 parent categories and 159 object types under a unified detection protocol.

The taxonomy is organized as a multi-level tree. The top level contains 30 common remote sensing categories covering transportation, energy facilities, public facilities, sports fields, maritime infrastructure, aerial platforms, and other objects. For object families where fine-grained recognition is visually meaningful, we further define detailed branches, including 45 aircraft types, 13 vehicle types, and 71 ship types. The dataset contains 702,707 fine-grained annotations, including 102,076 aircraft instances, 414,405 vehicle instances, and 186,226 ship instances, as shown in Fig. \ref{fig:ModelOverview}. These three branches make LEVIRDet-159 a multi-family fine-grained detection benchmark rather than a collection of independent family-specific datasets.

\subsubsection{Density, Category Breadth, and Object-Size Coverage}

Fig.~\ref{fig:combined_dataset_statistics} analyzes LEVIRDet-159 from three complementary views. The image-by-object-number distribution shows that the dataset contains both sparse scenes and dense scenes, instead of being dominated by either isolated large objects or crowd-like counting images. The image-by-category-count distribution further shows that the dataset includes both single-category scenes and multi-category scenes, which is necessary for learning object co-occurrence and suppressing category shortcuts in complex remote sensing imagery.

The object-size distribution is especially important for remote sensing detection. Using the square root of box area as the object-size measure, the positive-tile protocol contains 640,998 objects below 16 pixels and 946,768 objects in \([16,32)\). In proportion, 25.00\% of instances are smaller than 16 pixels and 61.93\% are smaller than 32 pixels, while 38.07\% are no smaller than 32 pixels. LEVIRDet-159 therefore covers both tiny-object detection and large-structure localization under one unified protocol, which is difficult to obtain from specialized benchmarks alone.

\subsubsection{Source and Sensor Diversity}

The images in LEVIRDet-159 are collected from a broad range of existing resources, including object detection (Levir-ship\cite{levirship}, DOTA\cite{dota}, UCAS-AOD\cite{ucasaod}, NWPU\cite{NWPU}, RSOD\cite{rsod}, HRRSD\cite{hrssd}, TAS\cite{tas}, DIOR\cite{DIOR}, HRSC\cite{HRSC2016}, VEDAI\cite{vedai}, ITCVD\cite{ITCVD}, DLR3K\cite{DLR3kDLR}, Bridge\cite{bridge}, OIRDS\cite{OIRDS}, MAR20\cite{mar20}, FAIR1M\cite{fair1m}, VHRShips\cite{vhrships}, ShipRSImageNet\cite{shiprsimagenet}, Ningbo\cite{Ningbo}, HRPlanev2\cite{hrplane}), segmentation (Vaihingen\cite{Vaihingen}, Potsdam\cite{Vaihingen}, Toronto\cite{Vaihingen}), classification (MTARSI\cite{MTARSI}), and pre-training datasets (fMoW\cite{FMOW}), as well as images released in public challenges (MASATI\cite{MASATI}, Airbus-ship\cite{AirbusShip}, Airbus-aircraft\cite{AirbusAircraft}).


The source distribution in Fig.~\ref{fig:data_source_donut_times_larger} shows that LEVIRDet-159 is not tied to a single imaging platform. Google/Baidu map-service imagery contributes 51.84\% of the images, while the remaining 48.16\% comes from other sources and sensors. Named satellite sensors account for a large portion of this diversity, including SPOT (18.38\%), GF-2 (11.89\%), JL-1 (1.89\%), GF-1 (1.44\%), GF-6 (1.03\%), and Pléiades (0.59\%). Aerial imagery contributes 5.09\%, Bing contributes 4.02\%, and other public sources contribute 3.84\%. This composition exposes the detector to different resolutions, imaging styles, regions, and acquisition conditions, which is essential for cross-dataset transfer.

Cross-source integration also introduces a risk. A large dataset may appear diverse while allowing the model to learn source-specific appearance or annotation conventions\cite{GLHTPAMI}. LEVIRDet-159 controls this risk at both the split and annotation levels. When an original source provides official partitions, we preserve them rather than randomly redistributing images after aggregation. For sources derived from large scenes or adjacent crops, splitting is performed at the parent-scene or large-image level so that neighboring or overlapping tiles do not appear in both training and evaluation. Images converted from segmentation, classification, or tracking resources inherit the split of the original image. During fine-grained relabeling, source identity is hidden from annotators, and the unified category scope is applied to all sources rather than only to newly introduced classes.

\subsubsection{Annotation and Protocol Unification}

\begin{figure}
\centering
\includegraphics[width=1\linewidth]{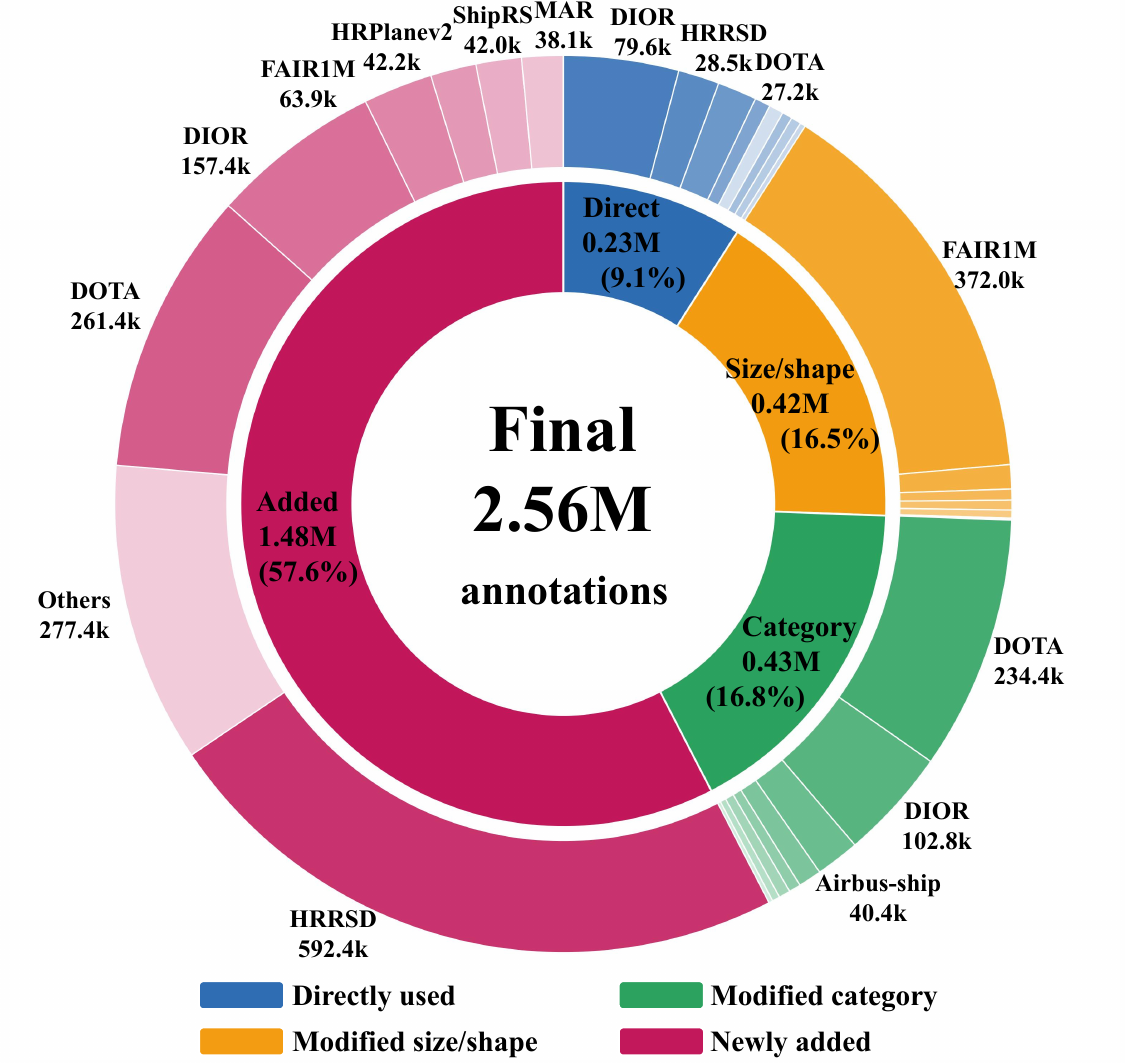}
\vspace{-4ex}
\caption{
 An annotation refinement composition of LEVIRDet-159. The nested chart separates directly inherited boxes, fine-grained relabeled boxes, geometry-corrected boxes, and newly added boxes, showing that the dataset is constructed through systematic annotation revision rather than simple dataset combination.
}
\label{fig:nested_donut_annotation_sources_larger_times}
\vspace{-4ex}
\end{figure}

LEVIRDet-159 is not a direct combination of public annotations. As shown in Fig.~\ref{fig:nested_donut_annotation_sources_larger_times}, only a small portion of boxes are directly inherited after protocol checking. Under the positive-tile protocol, 231,960 boxes are directly reused, 431,348 boxes receive fine-grained relabeling, 422,634 boxes are geometrically corrected, and 1,478,031 boxes are newly added. In other words, more than 90\% of the final positive-tile annotations are newly created or revised in some form.

This decomposition is important for interpreting the large number of newly added boxes. We do not treat the source datasets as inaccurate. Many were designed for different evaluation protocols or category scopes. New boxes mainly come from newly introduced categories, instances outside the original category scope, and all-category image-level review after taxonomy harmonization. The geometry-corrected boxes mainly arise from differences between original box styles and the tight horizontal-box protocol. For example, directly enclosing an oriented box can be much looser than the minimum tight horizontal box for a diagonally oriented aircraft or ship, which may substantially reduce IoU under COCO-style evaluation\cite{OBBTPAMI}. The fine-grained relabeled boxes correspond to existing localized objects whose parent labels are preserved while aircraft, vehicle, or ship subclasses are assigned when visual evidence is sufficient.

Fig.~\ref{fig:dataset_category_before_after_with_size_and_curve} further compares category coverage before and after unification. The expansion is not limited to adding a few rare classes. It increases coverage across common remote sensing parents, newly introduced scene objects, and fine-grained aircraft, vehicle, and ship branches. This makes the dataset useful for training a single detector with a broad category vocabulary, while preserving a strict localization standard.

\begin{figure*}
\centering
\includegraphics[width=0.9\linewidth]{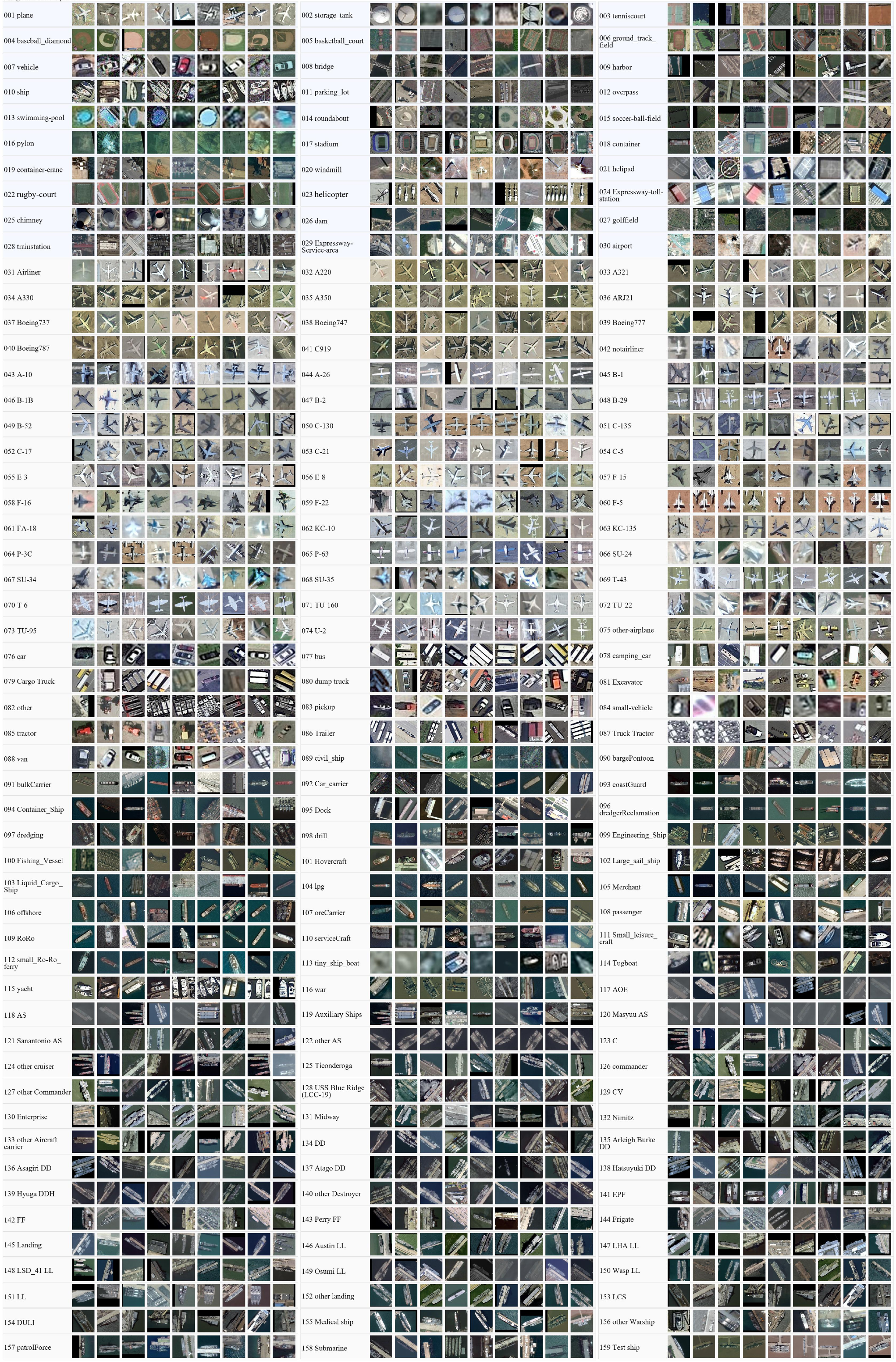}
\vspace{-1ex}
\caption{
A category-wise visualization of LEVIRDet-159, showing representative object crops across all categories.}
\label{fig:class_contact_sheet}
\end{figure*}
\subsubsection{Fine-Grained Hierarchy and Visual Evidence}

The examples in Fig.~\ref{fig:class_contact_sheet} illustrate why LEVIRDet-159 uses a hierarchy instead of forcing every instance into a flat leaf class. Aircraft, vehicles, and ships can often be distinguished at fine granularity when the object crop is sufficiently resolved, but their fine-grained identities may become ambiguous at coarse spatial resolution or under severe image degradation. LEVIRDet-159 therefore annotates each instance at the most specific reliable level supported by the image evidence and falls back to an ancestor category when fine-grained cues are not visually resolvable.

This hierarchy is not only a visual description of category relationships. It is used as an operational structure throughout the dataset and model. It defines valid mixed-depth labels during annotation, provides ancestor-path supervision during training, guides compatible matching, and supports tree-collapsed predictions.

\subsubsection{License, Privacy, and Release Plan}

LEVIRDet-159 is constructed from publicly accessible remote sensing resources, public challenge datasets, and map-service imagery under terms compatible with research use and dataset release. For transparency, the released dataset include a source manifest for every image tile, recording the source dataset or provider, sensor or platform when available, original split information, citation, license or usage-term category, and redistribution status. Source-specific attribution and usage requirements are preserved in the release documentation.

The dataset contains overhead remote sensing imagery of outdoor scenes and does not collect private user data, street-level imagery, faces, license plates, or individual-level behavioral information. The annotations describe object categories and bounding boxes rather than personal identities. The public release will include the full image tiles, annotations, metadata, preprocessing scripts, trained models, and evaluation code in a versioned repository accompanying the final paper. A contact and removal procedure will also be provided for source owners if attribution, access, or redistribution conditions change.

\subsection{Resolution-Aware Data Engine}
\label{sec:data_engine}

Building a universal remote sensing detection dataset requires more than collecting images. Source datasets differ in category scope, localization protocol, and semantic granularity. If these differences are not resolved, a detector may learn source-specific annotation styles rather than transferable object semantics. We therefore construct LEVIRDet-159 with a resolution-aware data engine that separates localization standardization, fine-grained relabeling, and full-image consistency review.

The data engine follows two principles. First, box geometry is standardized before fine-grained recognition, so all objects follow the tight horizontal bounding box protocol. Second, semantic depth is determined by visual evidence. When fine-grained identity is unreliable because of low resolution or image degradation, the annotation falls back to the most reliable ancestor category instead of forcing a noisy leaf label. This produces dense localization supervision while recording the limits of semantic observability under different GSDs.



\subsubsection{Stage I: Geometry-First Parent Annotation}

The first stage establishes a unified localization protocol. Annotators inspect each image under the full LEVIRDet-159 category scope and localize every supported object with the tightest horizontal rectangle around its visible extent. Only parent categories are assigned at this stage, which prevents uncertain fine-grained recognition from affecting box drawing.


This stage also converts heterogeneous source annotations into the unified tight-HBB protocol. Existing horizontal boxes are checked for tightness, while oriented boxes are treated as candidates rather than directly converted into enclosing rectangles~\cite{OBBTPAMI,OBB2TPAMI}. For elongated objects such as aircraft and ships, such conversion can produce loose boxes and reduce IoU under COCO-style evaluation. We adopt HBBs to first study unified, general-purpose detection under a common annotation protocol. Extending this protocol to OBBs requires consistent geometry and orientation definitions, including rules for objects without a clear principal axis.

To improve geometric quality control, we use segmentation-assisted review as an outlier detector. For each candidate object, a segmentation model produces a mask proposal whose minimum enclosing rectangle is compared with the human box. If the side-length difference exceeds a tolerance or the IoU is below a threshold, the object is sent to an additional review round. The segmentation model is not used to replace human annotation. It only surfaces suspicious boxes. This stage produces directly inherited boxes, geometry-corrected boxes, and newly added parent boxes under one protocol.

\subsubsection{Stage II: Source-Blind Fine-Grained Relabeling}

After geometry is stabilized, parent-level boxes are cropped and sent to a source-blind labeling interface. Annotators label object crops without seeing the source dataset identity, reducing reliance on dataset-specific naming or appearance cues. Crops are presented from large to small so that annotators first establish reliable visual distinctions from clear examples and then handle smaller or more ambiguous instances. Clear instances are assigned fine-grained aircraft, vehicle, or ship labels, while ambiguous ones remain at the most reliable ancestor node.


This evidence-based fallback retains ambiguous objects as valid coarse-level samples, allowing mixed-granularity supervision without introducing noisy leaf labels.




\subsubsection{Stage III: Projection-Back Review and Global Calibration}

The third stage projects fine-grained labels back to the original image coordinates and reviews the results in full-image context. This step corrects errors that are difficult to judge from isolated crops, including duplicate or missing instances, context mismatches, and local density inconsistency. All images undergo multiple rounds of calibration so that box tightness, label granularity, and instance completeness follow the same protocol.

Projection-back review is especially important for the large number of newly added boxes. These boxes are not produced by indiscriminately densifying every image or by only supplementing newly introduced categories. Instead, reviewers inspect all categories under the unified scope. For small objects, the local crop and surrounding scene are checked together to avoid over-dense or context-inconsistent annotation. If the object extent or fine-grained identity is not visually supportable, the annotation is either rejected or kept at a reliable parent level.

After the three stages, each annotation belongs to one of four interpretable outcomes: directly inherited, geometry corrected, fine-grained relabeled, or newly added. This decomposition explains why LEVIRDet-159 contains many new and revised boxes without implying that source datasets were wrong. They were often built for different evaluation protocols and category scopes. In addition, public imagery with available resolution metadata is collected to train the online GSD predictor in LEVIRDetNet, linking the resolution-aware annotation protocol with the scale-aware model design.

\section{Methodology}


\subsection{LEVIRDetNet Model Framework}

\begin{figure*}
\centering
\includegraphics[width=1\linewidth]{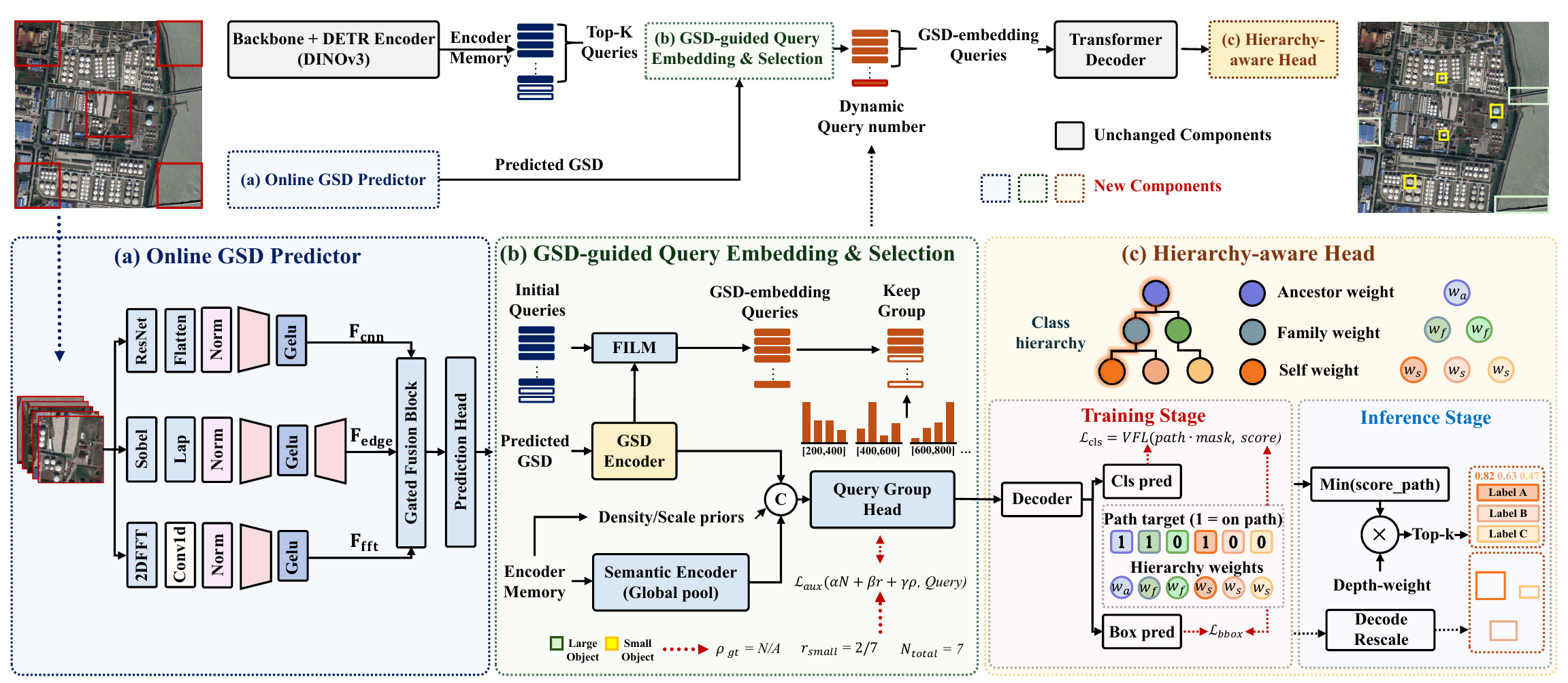}
\vspace{-2ex}
\caption{
An overview of the proposed LEVIRDetNet. The framework adapts a DEIM-style end-to-end detector to universal remote sensing object detection by integrating visual scale estimation, scene-aware query allocation, and hierarchy-aware classification under the LEVIRDet-159 annotation protocol. Panel (a) shows the online GSD predictor. Panel (b) illustrates how the predicted GSD is embedded into decoder queries and used for active query budget prediction. Panel (c) presents the hierarchy-aware detection head and summarizes the training and inference paths.} 
\label{fig:overall}
\vspace{-4ex}
\end{figure*}

LEVIRDetNet is a scale-hierarchy-aware detection foundation model built on a DEIM-style end-to-end detector with a DINOv3-initialized ViT backbone and transformer encoder and decoder, as illustrated in Fig. \ref{fig:overall}. The backbone, encoder, box branch, and distribution-refinement losses follow the base detector. All components are fully trained without LoRA. Three coupled modules connect visual scale, scene density, and semantic granularity: GSD-aware query conditioning modulates query embeddings using online visual GSD estimates, dynamic query allocation adapts query budgets to scene density and scale, and a hierarchy-aware head learns from mixed-granularity labels while supporting flat benchmark evaluation. These modules strengthen the detector under large scale variation, dense scenes, and hierarchical category systems.

\subsection{GSD-Aware Query Conditioning}

The apparent size of objects in overhead imagery is strongly affected by ground sampling distance (GSD). Two images resized to the same input resolution may correspond to different physical footprints, and the same pixel-sized object may provide different semantic evidence under different GSDs. Instead of relying on external metadata, which may be unavailable or inconsistent, we learn an online visual GSD cue and use it as a detector-conditioning signal, as illustrated in Fig. \ref{fig:overall} (a). 

For each image, the GSD branch extracts \(K\) fixed patches \(\{P_k\}_{k=1}^{K}\). For large images, patches are sampled from the four corners and the image center. For smaller images, center-biased patches are used. Each patch is normalized and encoded by three complementary descriptors:
\begin{equation}
e_k^{\mathrm{cnn}} = f_{\mathrm{cnn}}(P_k), \quad
e_k^{\mathrm{fft}} = f_{\mathrm{fft}}(P_k), \quad
e_k^{\mathrm{edge}} = f_{\mathrm{edge}}(P_k).
\end{equation}
where \(P_k\) denotes the \(k\)-th sampled image patch, and \(K\) is the number of sampled patches. \(f_{\mathrm{cnn}}(\cdot)\), \(f_{\mathrm{fft}}(\cdot)\), and \(f_{\mathrm{edge}}(\cdot)\) denote the CNN, frequency-domain, and edge-statistic descriptor extractors, respectively. \(e_k^{\mathrm{cnn}}\), \(e_k^{\mathrm{fft}}\), and \(e_k^{\mathrm{edge}}\) are the corresponding descriptors of patch \(P_k\).

The CNN descriptor captures semantic and texture cues. The FFT descriptor summarizes radial frequency statistics from the gray-scale spectrum. The edge descriptor encodes Sobel and Laplacian statistics, which are useful for estimating spatial detail and image sharpness.

The three descriptors are fused by a learned gate:
\begin{equation}
\alpha_k =
\mathrm{softmax}
\left(
W_g
\left[
e_k^{\mathrm{cnn}},
e_k^{\mathrm{fft}},
e_k^{\mathrm{edge}}
\right]
\right),
\end{equation}
where \(\alpha_k\) denotes the descriptor-fusion weight vector of patch \(P_k\). \(W_g\) is the learnable gating projection. \([\cdot]\) denotes feature concatenation, and \(\mathrm{softmax}(\cdot)\) normalizes the descriptor weights across the three descriptor types.

\begin{equation}
e_k =
\alpha_{k,1} e_k^{\mathrm{cnn}} +
\alpha_{k,2} e_k^{\mathrm{fft}} +
\alpha_{k,3} e_k^{\mathrm{edge}} .
\end{equation}
where \(e_k\) is the fused representation of patch \(P_k\). \(\alpha_{k,1}\), \(\alpha_{k,2}\), and \(\alpha_{k,3}\) are the three components of \(\alpha_k\), corresponding to the CNN, FFT, and edge descriptors.

Patch-level features are then aggregated by attention:
\begin{equation}
a_k = \mathrm{softmax}(w_a^\top e_k), 
\qquad
e_I = \sum_{k=1}^{K} a_k e_k .
\end{equation}
where \(a_k\) denotes the attention weight of patch \(P_k\). \(w_a\) is the learnable attention vector. \(e_I\) is the image-level representation aggregated from all sampled patches.

A regression head predicts the image-level GSD estimate \(\hat g\). To reduce the dynamic range, we use the logarithmic GSD value:
\begin{equation}
z_g = \psi\left(\log(\max(\hat g,\epsilon))\right),
\end{equation}
where \(\hat g\) denotes the predicted visual GSD. \(\epsilon\) is a small constant for numerical stability. \(\log(\cdot)\) compresses the dynamic range of the predicted GSD. \(\psi(\cdot)\) is a two-layer MLP followed by layer normalization, and \(z_g\) is the resulting GSD embedding.

The GSD embedding \(z_g\) modulates the initial decoder query encoding \(q_i^0\). In the Feature-wise Linear Modulation (FILM\cite{perez2018film}) mode used in our implementation, \(z_g\) is mapped to a scale-shift pair \((\gamma_g,\beta_g)\), and each query is updated as
\begin{equation}
\tilde q_i^0 =
q_i^0 \odot (1 + \lambda \gamma_g) + \lambda \beta_g ,
\end{equation}
where \(q_i^0\) denotes the original initial embedding of query \(i\), and \(\tilde q_i^0\) is the GSD-modulated query embedding. \(\gamma_g\) and \(\beta_g\) denote the GSD-conditioned scale and shift terms, respectively. \(\lambda\) controls the modulation strength, and \(\odot\) denotes element-wise multiplication.

This design does not impose hard scale rules. Instead, it biases the decoder toward a resolution-aware interpretation of candidate objects.

\subsection{Dynamic Query Allocation}

Remote sensing images range from sparse large-object scenes to dense small-object regions. A fixed query budget is therefore inefficient. Too few queries may miss objects in dense scenes, while too many queries introduce redundant decoding and matching cost in sparse scenes. We introduce a dynamic query allocation module that predicts an image-specific active query budget while preserving the ranked query structure of the base detector, as illustrated in Fig. \ref{fig:overall} (b). 

The module receives four types of information: the global encoder memory \(\bar M\), the GSD embedding \(z_g\), a density prior \(p_{\mathrm{density}}\), and a scale prior \(p_{\mathrm{scale}}\). The density prior is computed from the average channel variance of encoder memory, while the scale prior measures the energy contrast between coarse and fine encoder levels. These visual priors describe the resized feature distribution, while \(z_g\) provides complementary physical-scale information.

The input to the query allocator is
\begin{equation}
u =
\left[
\bar M,
z_g,
p_{\mathrm{density}},
p_{\mathrm{scale}}
\right].
\end{equation}
where \(u\) denotes the input feature of the query allocator. \(\bar M\) is the global encoder-memory representation. \(z_g\) is the GSD embedding. \(p_{\mathrm{density}}\) and \(p_{\mathrm{scale}}\) denote the density prior and scale prior, respectively. \([\cdot]\) denotes concatenation.

The allocator predicts logits over a discrete query budget set
\(\mathcal{B}=\{300,600,800\}\):
\begin{equation}
\pi = h_q(u), 
\qquad
\hat b = \arg\max_{b\in\mathcal{B}} \pi_b .
\end{equation}
where \(h_q(\cdot)\) denotes the query-budget prediction head. \(\pi\) is the predicted logit vector over the candidate budget set \(\mathcal{B}\). \(\pi_b\) is the logit corresponding to budget \(b\), and \(\hat b\) is the predicted active query budget.

During training, the dynamic query module is supervised without changing the main Hungarian assignment and denoising layout. The target budget is derived from the number of ground-truth objects, the fraction of small objects, the normalized object density, and a scale bonus:
\begin{equation}
s =
w_n n_{\mathrm{gt}} +
w_s r_{\mathrm{small}} +
w_d d +
w_b \max(0, r_0-\bar r),
\end{equation}
where \(s\) denotes the continuous scene-complexity score used to derive the target query budget. \(n_{\mathrm{gt}}\) is the number of ground-truth objects, \(r_{\mathrm{small}}\) is the small-object ratio, \(d\) is the normalized object density, and \(\bar r\) is the mean normalized object scale. \(w_n\), \(w_s\), \(w_d\), and \(w_b\) are scalar weights. \(r_0\) is the threshold used by the scale bonus, and \(\max(\cdot)\) keeps the bonus active only when the mean object scale is smaller than \(r_0\).

The target budget \(b^\star\) is assigned to the nearest bin in \(\mathcal{B}\), and the auxiliary loss is
\begin{equation}
\mathcal{L}_{\mathrm{dyn}}
=
\lambda_q \mathrm{CE}(\pi,b^\star).
\end{equation}
where \(b^\star\) denotes the target query budget. \(\mathcal{L}_{\mathrm{dyn}}\) is the auxiliary dynamic query loss. \(\lambda_q\) is the loss weight, and \(\mathrm{CE}(\cdot)\) denotes the cross-entropy loss between the predicted budget logits \(\pi\) and the target budget \(b^\star\).

At inference, the predicted budget selects the active prefix of ranked queries, while inactive queries are masked out. This allows the detector to retain high capacity for dense small-object scenes and reduce redundant decoding for sparse scenes.

\subsection{Hierarchy-Aware Detection Head}

LEVIRDet-159 contains both coarse parent categories and fine-grained subclasses. Its taxonomy includes 30 common remote sensing parent categories and fine-grained branches for aircraft, vehicles, and ships. This structure allows the annotation protocol to assign the most specific visually supported label and fall back to an ancestor category when fine-grained evidence is insufficient, as illustrated in Fig. \ref{fig:overall} (c). 

A flat classifier is not ideal for such mixed-granularity supervision, because it may penalize predictions that are semantically compatible with the annotation but appear at a different taxonomy depth. We therefore replace the flat classification head with a hierarchy-aware multi-label head. The box regression branch is kept the same as the base detector, while the classification branch produces sigmoid scores over all taxonomy nodes.

Let \(y\) be the ground-truth label of a matched object, and let \(\mathcal{A}(y)\) denote the set containing \(y\) and all its ancestors. For a predicted box with IoU quality \(r\), the classification target for class \(c\) is defined as
\begin{equation}
t_c(y) =
\begin{cases}
r, & c \in \mathcal{A}(y),\\
0, & \mathrm{otherwise}.
\end{cases}
\end{equation}
where \(t_c(y)\) denotes the target score for class \(c\) under ground-truth label \(y\). \(y\) is the ground-truth taxonomy node, \(c\) is a candidate taxonomy node, and \(\mathcal{A}(y)\) is the ancestor path of \(y\) including \(y\) itself. \(r\) denotes the IoU quality of the matched predicted box.

The target is optimized with a varifocal-style sigmoid loss.

To reduce harsh penalties between semantically related categories, we introduce a class-pair weight matrix \(W\):
\begin{equation}
W_{y,c} =
\begin{cases}
1, & c \in \mathcal{A}(y),\\
0, & c \in \mathcal{D}(y),\\
\lambda_{\mathrm{sib}}, & c \in \mathcal{S}(y),\\
\lambda_{\mathrm{fam}}, & c \in \mathcal{F}(y),\\
1, & \mathrm{otherwise},
\end{cases}
\end{equation}
where \(W_{y,c}\) denotes the class-pair weight between ground-truth label \(y\) and class \(c\). \(\mathcal{D}(y)\), \(\mathcal{S}(y)\), and \(\mathcal{F}(y)\) denote the descendant, sibling, and same-family sets of class \(y\), respectively. \(\lambda_{\mathrm{sib}}\) and \(\lambda_{\mathrm{fam}}\) are scalar weights for sibling and same-family negatives.

The hierarchy-aware classification loss is
\begin{equation}
\mathcal{L}_{\mathrm{hcls}}
=
\sum_{c} W_{y,c}
\mathcal{L}_{\mathrm{vfl}}(p_c,t_c(y)).
\end{equation}
where \(\mathcal{L}_{\mathrm{hcls}}\) denotes the hierarchy-aware classification loss. \(p_c\) is the predicted score of class \(c\), and \(\mathcal{L}_{\mathrm{vfl}}(\cdot)\) denotes the varifocal loss.

The same multi-level targets are also used for denoising queries, ensuring that the auxiliary denoising objective is consistent with the main detection loss.

\subsection{Multi-Level Matching and Tree-Collapsed Prediction}

LEVIRDetNet follows the end-to-end assignment paradigm of DETR-like detectors. In the early training stage, we use a hierarchy-compatible Hungarian assigner with three costs: classification cost, \(L_1\) box cost, and generalized IoU cost. The classification term considers labels compatible with the ground-truth path and prevents coarse and fine labels along the same taxonomy path from competing during matching.

After the detector becomes stable, the matching objective is switched to a DEIM-style quality-aware cost. For a ground-truth label \(y\), the hierarchy-compatible class evidence is computed as
\begin{equation}
\hat p_y =
\max_{c}
\left(
p_c C_{y,c}
\right),
\end{equation}
where \(\hat p_y\) denotes the hierarchy-compatible class evidence for ground-truth label \(y\). \(p_c\) is the predicted score of class \(c\), and \(C_{y,c}\) is the taxonomy compatibility mask between ground-truth label \(y\) and class \(c\). \(\max_c(\cdot)\) selects the strongest compatible class evidence.

The switched matching cost is
\begin{equation}
\mathcal{C}_{\mathrm{switch}}
=
-\hat p_y \cdot
\mathrm{IoU}(b,\hat b)^{\alpha}.
\end{equation}
where \(\mathcal{C}_{\mathrm{switch}}\) denotes the switched matching cost. \(b\) and \(\hat b\) denote the ground-truth and predicted boxes, respectively. \(\mathrm{IoU}(\cdot)\) is the intersection-over-union function, and \(\alpha\) controls the strength of IoU-quality weighting.

This cost emphasizes high-quality localization and compatible semantic confidence in the late stage of training.

At inference, hierarchical predictions are collapsed to one label per query for standard benchmark evaluation. We first replace the score of each class by the minimum score along its ancestor path:
\begin{equation}
s_c^{\mathrm{path}}
=
\min_{a\in\mathcal{A}(c)} p_a .
\end{equation}
where \(s_c^{\mathrm{path}}\) denotes the path-consistent score of class \(c\). \(\mathcal{A}(c)\) is the ancestor path of \(c\) including \(c\) itself. \(a\) denotes an ancestor node on this path, and \(p_a\) is the predicted score of node \(a\). \(\min(\cdot)\) requires all nodes on the path to be consistently confident.

Then each query selects the class with the highest depth-adjusted score:
\begin{equation}
\hat c =
\arg\max_c
s_c^{\mathrm{path}}
\left(
1+\rho(\mathrm{depth}(c)-1)
\right),
\end{equation}
where \(\hat c\) denotes the final collapsed class label. \(s_c^{\mathrm{path}}\) is the path-consistent score of class \(c\). \(\mathrm{depth}(c)\) is the taxonomy depth of class \(c\), and \(\rho\) is a small coefficient that mildly favors confident fine-grained labels. The selected predictions are decoded into bounding boxes.

\subsection{Implementation Details}

\subsubsection{Network Design}

LEVIRDetNet is implemented using MMDetection \cite{chen2019mmdetection} and MMEngine \cite{mmengine2022} frameworks. The default input resolution is \(1024\times1024\), and batch-level random resizing samples from \(\{896,960,1024,1088,1152\}\). The detector uses 800 object queries before dynamic query allocation, and the query budget bins are \(\{300,600,800\}\). The GSD branch predicts from five \(224\times224\) patches using a ResNet-50 CNN encoder, 64 radial FFT bins, gated descriptor fusion, and attention-based patch aggregation. GSD FILM modulation uses \(\lambda=0.5\), and the dynamic query auxiliary loss weight is set to 0.05. For the hierarchy-aware head, the descendant negative weight is 0, the sibling negative weight is 0.25, and the same-family negative weight is 0.5.

\subsubsection{Training Details}

Following the model's structural design, we adopt a two-stage training strategy. 

First, the online GSD predictor is pretrained on an auxiliary GSD-labeled remote sensing dataset containing approximately 800k \(512\times512\) image patches. The images are collected from public remote sensing resources with available spatial-resolution metadata, including ISPRS\cite{Vaihingen}, VEDAI\cite{vedai}, COWC\cite{cowc}, DLR\cite{DLR3kDLR}, Inria Aerial\cite{iail}, xView\cite{xview}, fMoW\cite{FMOW}, and BigEarth\cite{bigearth}. Each patch is assigned a GSD label in meters per pixel (m/pixel). This auxiliary GSD pretraining set, together with its source metadata and license manifest, will be released as an accompanying component of LEVIRDet-159. This stage is trained for 1.6M iterations, enabling the predictor to learn a robust visual representation of spatial resolution from image content.

Second, the full LEVIRDetNet is trained on LEVIRDet-159 for 120 epochs. The pretrained GSD predictor is loaded into the detector but is not frozen. It is jointly optimized with the detection network so that the GSD representation remains physically meaningful while adapting to query modulation, dynamic query allocation, and hierarchy-aware detection. For optimization, we employ the AdamW optimizer with an initial learning rate of $5\times10^{-4}$ and a weight decay of $1.25\times10^{-4}$. The learning rate is adjusted using a cosine annealing scheduler with linear warm-up. Training is conducted using bfloat16 precision.

\begin{table*}[!t]
\caption{Comparison with other methods on ADCOS~\cite{adcos}, UCAS-AOD~\cite{ucasaod}, HRPlane-v2~\cite{hrplane}, and CORS-ADD~\cite{corsadd}.}
\label{tab:comparisons_aircraft}
\centering
\resizebox{\textwidth}{!}{
\begin{tabular}{c|c c c|c c c|c c c|c c c}
\toprule
\multirow{2}{*}{\shortstack[c]{Method}} &
\multicolumn{3}{c|}{ADCOS~\cite{adcos}} &
\multicolumn{3}{c|}{UCAS-AOD~\cite{ucasaod}} &
\multicolumn{3}{c|}{HRPlane-v2~\cite{hrplane}} &
\multicolumn{3}{c}{CORS-ADD~\cite{corsadd}} \\
\cline{2-13}
&
\rule{0pt}{3.0ex}$\text{AP}_{\text{bbox}}$ &
$\text{AP}^{50}_{\text{bbox}}$ &
$\text{AP}^{75}_{\text{bbox}}$ &
$\text{AP}_{\text{bbox}}$ &
$\text{AP}^{50}_{\text{bbox}}$ &
$\text{AP}^{75}_{\text{bbox}}$ &
$\text{AP}_{\text{bbox}}$ &
$\text{AP}^{50}_{\text{bbox}}$ &
$\text{AP}^{75}_{\text{bbox}}$ &
$\text{AP}_{\text{bbox}}$ &
$\text{AP}^{50}_{\text{bbox}}$ &
$\text{AP}^{75}_{\text{bbox}}$ \\
\midrule
HrNet~\cite{hrnet} & 76.73 & 92.03 & 90.50 & 69.32 & 89.55 & 81.43 & 75.56 & 97.72 & 92.66 & 63.19 & 89.61 & 72.68 \\
Fast R-CNN~\cite{fastrcnn} & 70.42 & 90.89 & 84.31 & 68.08 & 90.68 & 80.66 & 69.25 & 96.45 & 86.72 & 60.79 & 89.19 & 69.75 \\
YOLOv3~\cite{yolo} & 54.51 & 85.64 & 61.36 & 51.04 & 86.27 & 53.96 & 65.39 & 96.03 & 75.18 & 52.17 & 85.80 & 56.89 \\
YOLOF~\cite{yolof} & 47.14 & 74.16 & 55.05 & 37.29 & 74.16 & 34.54 & 61.58 & 94.80 & 75.66 & 43.69 & 76.05 & 45.66 \\
Sparse R-CNN~\cite{sparse_rcnn} & 58.70 & 74.45 & 69.80 & 73.67 & 88.91 & 86.98 & 48.02 & 69.62 & 58.75 & 51.79 & 76.90 & 58.21 \\
RTMDet~\cite{rtmdet} & \cellcolor{gray}79.45 & 91.87 & 89.62 & 74.32 & \cellcolor{lightgray}94.40 & 84.94 & 77.37 & 98.50 & 93.67 & \cellcolor{lightgray}69.40 & 92.78 & \cellcolor{lightgray}80.58 \\
EfficientDet~\cite{efficientdet} & 68.95 & 90.77 & 82.27 & 64.88 & 91.58 & 75.95 & 65.00 & 96.81 & 80.03 & 62.77 & 91.10 & 71.76 \\
CenterNet~\cite{centernet} & 59.75 & 89.90 & 69.75 & 41.47 & 81.36 & 41.56 & 45.55 & 67.87 & 52.93 & 43.55 & 82.08 & 41.62 \\
Deformable DETR~\cite{deformable_detr} & 60.56 & 86.40 & 71.93 & 60.87 & 88.78 & 71.84 & 70.29 & 95.67 & 87.38 & 58.16 & 88.78 & 66.06 \\
DINO~\cite{dino} & 70.72 & 96.17 & 87.71 & 66.71 & 89.80 & 78.04 & 76.09 & 97.59 & 93.13 & 63.54 & 92.69 & 74.72 \\
FCOS~\cite{fcos} & 66.83 & 89.81 & 79.76 & 37.00 & 86.64 & 18.89 & 71.04 & 96.04 & 89.27 & 59.19 & 89.66 & 65.47 \\
RetinaNet~\cite{retinanet} & 53.25 & 80.46 & 62.92 & 64.88 & 91.58 & 75.95 & 58.66 & 92.62 & 70.78 & 49.15 & 80.58 & 53.10 \\
TridentNet~\cite{tridentnet} & 70.23 & 89.83 & 82.26 & 61.50 & 89.12 & 72.32 & 72.50 & 96.52 & 89.42 & 61.58 & 88.05 & 69.81 \\
SSD~\cite{ssd} & 70.53 & 95.50 & 85.37 & 58.23 & 89.62 & 67.64 & 68.14 & 97.59 & 85.00 & 57.58 & 89.13 & 65.84 \\
Faster R-CNN~\cite{FasterR-CNN} & 72.39 & \cellcolor{darkgray}98.92 & 89.65 & 67.71 & 87.48 & 80.02 & 73.08 & 97.64 & 91.77 & 59.84 & 87.94 & 67.86 \\
YOLOX~\cite{yolox} & 77.96 & 98.13 & \cellcolor{lightgray}94.48 & 68.86 & 93.27 & 80.68 & 73.44 & \cellcolor{lightgray}98.51 & 91.74 & 67.99 & 94.25 & 79.64 \\
YOLOv12x~\cite{yolov12} & \cellcolor{lightgray}78.72 & 98.67 & \cellcolor{gray}95.31 & \cellcolor{lightgray}74.67 & 94.34 & \cellcolor{lightgray}86.31 & \cellcolor{lightgray}78.51 & \cellcolor{darkgray}98.59 & \cellcolor{lightgray}94.83 & \cellcolor{gray}71.70 & \cellcolor{darkgray}95.46 & \cellcolor{gray}82.69 \\
YOLOv10s-C2F-SIoU~\cite{yolosiou} & -- & -- & -- & -- & -- & -- & -- & 92.20 & -- & -- & -- & -- \\
YOLOv10s-C2F~\cite{yolosiou} & -- & -- & -- & -- & -- & -- & -- & 96.50 & -- & -- & -- & -- \\
DEIM~\cite{deim} & 77.46 & 97.77 & 94.46 & 69.37 & 92.08 & 81.05 & 71.66 & 97.54 & 90.73 & 67.08 & 94.30 & 78.76 \\
DEIMv2 (DINOv3 + CNN)~\cite{deimv2} & 77.54 & 98.51 & 94.30 & 54.08 & 85.46 & 61.30 & 74.81 & 97.87 & 90.84 & 67.39 & 94.27 & 78.14 \\
DEIMv2 (DINOv3 + ViT-Base)~\cite{deimv2} & 77.79 & \cellcolor{lightgray}98.68 & 84.46 & \cellcolor{gray}75.38 & \cellcolor{gray}96.14 & \cellcolor{gray}87.40 & \cellcolor{gray}78.87 & 98.47 & \cellcolor{gray}95.17 & 68.26 & \cellcolor{gray}95.28 & 79.41 \\
DynamicVis-base~\cite{dynamicvis} & 76.10 & 91.70 & 89.00 & 69.30 & 93.60 & 80.80 & 72.50 & 96.20 & 89.40 & 65.00 & 89.40 & 74.20 \\
DynamicVis-large~\cite{dynamicvis} & 77.10 & 91.80 & 89.50 & 69.90 & 93.60 & 82.00 & 75.00 & 97.30 & 91.70 & 64.60 & 90.90 & 74.50 \\
\midrule
\textbf{LEVIRDetNet (+4.15, +5.02, +3.05, +0.36)} &
\cellcolor{darkgray}\textbf{83.60} & \cellcolor{gray}\textbf{98.82} & \cellcolor{darkgray}\textbf{97.55} &
\cellcolor{darkgray}\textbf{80.40} & \cellcolor{darkgray}\textbf{98.63} & \cellcolor{darkgray}\textbf{90.62} &
\cellcolor{darkgray}\textbf{81.92} & \cellcolor{gray}\textbf{98.53} & \cellcolor{darkgray}\textbf{96.10} &
\cellcolor{darkgray}\textbf{72.06} & \cellcolor{lightgray}\textbf{94.88} & \cellcolor{darkgray}\textbf{83.56} \\
\bottomrule
\end{tabular}}
\end{table*}

\begin{table*}[!t]
\caption{Comparison with other methods on SkyFusion-plane~\cite{skyfusion}, VHRV\cite{vhrv}, SkyFusion-ship~\cite{skyfusion}, and NWPU~\cite{NWPU}.}
\label{tab:comparisons_plane_ship_general}
\centering
\resizebox{\textwidth}{!}{
\begin{tabular}{c|c c c|c c c|c c c|c c c}
\toprule
\multirow{2}{*}{\shortstack[c]{Method}} &
\multicolumn{3}{c|}{SkyFusion-plane~\cite{skyfusion}} &
\multicolumn{3}{c|}{VHRV} &
\multicolumn{3}{c|}{SkyFusion-ship~\cite{skyfusion}} &
\multicolumn{3}{c}{NWPU~\cite{NWPU}} \\
\cline{2-13}
&
\rule{0pt}{3.0ex}$\text{AP}^{50}_{\text{bbox}}$ &
$\text{AP}^{s}_{\text{bbox}}$ &
$\text{AP}^{m}_{\text{bbox}}$ &
$\text{AP}_{\text{bbox}}$ &
$\text{AP}^{50}_{\text{bbox}}$ &
$\text{AP}^{75}_{\text{bbox}}$ &
$\text{AP}^{50}_{\text{bbox}}$ &
$\text{AP}^{s}_{\text{bbox}}$ &
$\text{AP}^{m}_{\text{bbox}}$ &
$\text{AP}_{\text{bbox}}$ &
$\text{AP}^{50}_{\text{bbox}}$ &
$\text{AP}^{75}_{\text{bbox}}$ \\
\midrule
HrNet~\cite{hrnet} & 90.64 & 59.15 & 69.47 & 65.05 & 82.81 & 77.14 & 31.85 & 10.94 & 47.32 & 67.81 & 91.77 & 79.23 \\
Fast R-CNN~\cite{fastrcnn} & 90.10 & 45.98 & 61.59 & 62.99 & 82.59 & 75.13 & 21.53 & 7.70 & \cellcolor{gray}49.57 & 60.14 & 89.77 & 70.11 \\
YOLOv3~\cite{yolo} & 91.28 & 49.13 & 65.27 & 16.06 & 42.64 & 9.18 & 25.07 & 6.38 & 34.73 & 49.80 & 85.08 & 53.43 \\
YOLOv5xu~\cite{yolov5} & \cellcolor{lightgray}96.30 & -- & -- & -- & -- & -- & \cellcolor{gray}47.70 & -- & -- & -- & -- & -- \\
YOLOF~\cite{yolof} & 83.38 & 26.47 & 54.30 & 8.73 & 24.78 & 5.04 & 11.32 & 2.46 & 45.92 & 50.69 & 85.78 & 54.49 \\
Sparse R-CNN~\cite{sparse_rcnn} & 51.27 & 20.81 & 28.28 & 46.27 & 59.57 & 53.96 & 31.30 & 9.59 & 28.78 & 52.13 & 74.53 & 57.02 \\
RTMDet~\cite{rtmdet} & 92.00 & 60.70 & 70.00 & 64.11 & 81.72 & 76.04 & 30.20 & 10.70 & 35.42 & 67.01 & 91.82 & 77.93 \\
EfficientDet~\cite{efficientdet} & 91.32 & 48.55 & 60.87 & 35.97 & 66.31 & 33.16 & 25.15 & 7.81 & 6.15 & 52.75 & 86.11 & 60.97 \\
Deformable DETR~\cite{deformable_detr} & 94.40 & 54.11 & 70.53 & 32.86 & 62.08 & 31.90 & 19.86 & 5.08 & 11.28 & 47.54 & 83.93 & 45.71 \\
DINO~\cite{dino} & 94.28 & 59.67 & 72.75 & 58.91 & 85.91 & 74.41 & 44.74 & 15.40 & 45.20 & 66.06 & 87.56 & 74.74 \\
FCOS~\cite{fcos} & 91.36 & 54.64 & 66.87 & 33.31 & 64.60 & 30.28 & 40.94 & 13.99 & 42.12 & 44.17 & 70.99 & 48.14 \\
RetinaNet~\cite{retinanet} & 66.48 & 13.26 & 35.98 & 32.64 & 62.24 & 30.43 & 25.15 & 7.81 & 6.15 & 22.43 & 43.41 & 21.04 \\
TridentNet~\cite{tridentnet} & 90.37 & 51.21 & 64.33 & 46.39 & 74.03 & 49.34 & 5.94 & 0.51 & 31.85 & 62.70 & 91.71 & 70.51 \\
SSD~\cite{ssd} & 89.66 & 46.56 & 62.23 & 37.22 & 69.58 & 39.19 & 28.77 & 7.62 & 36.18 & 59.86 & 91.25 & 65.85 \\
Faster R-CNN~\cite{FasterR-CNN} & 90.52 & 53.26 & 66.38 & 61.67 & 80.61 & 74.67 & 26.21 & 7.61 & 45.52 & 63.20 & 90.84 & 74.72 \\
YOLOX~\cite{yolox} & 91.47 & 56.40 & 67.04 & 54.81 & 86.65 & 67.87 & 35.64 & 4.29 & 9.16 & 65.81 & 92.02 & 75.41 \\
YOLOv12x~\cite{yolov12} & 94.52 & 61.72 & \cellcolor{gray}76.70 & \cellcolor{gray}72.58 & \cellcolor{gray}94.11 & \cellcolor{gray}84.04 & 43.56 & \cellcolor{gray}17.66 & 16.74 & 65.01 & 88.81 & 74.13 \\
DEIM~\cite{deim} & 91.05 & 50.39 & 67.27 & 54.88 & 79.13 & 66.38 & 17.90 & 5.71 & 15.62 & 53.81 & 77.09 & 63.07 \\
DEIMv2 (DINOv3 + CNN)~\cite{deimv2} & 91.47 & 53.90 & 68.94 & 18.44 & 38.05 & 18.14 & 27.58 & 8.11 & 28.92 & \cellcolor{lightgray}72.09 & \cellcolor{gray}94.70 & \cellcolor{lightgray}82.69 \\
DEIMv2 (DINOv3 + ViT-Base)~\cite{deimv2} & \cellcolor{gray}97.58 & \cellcolor{lightgray}64.62 & \cellcolor{darkgray}78.79 & \cellcolor{lightgray}67.58 & \cellcolor{lightgray}92.21 & \cellcolor{lightgray}82.14 & 44.10 & 15.18 & \cellcolor{lightgray}49.06 & \cellcolor{gray}73.60 & \cellcolor{darkgray}97.18 & \cellcolor{darkgray}86.40 \\
DynamicVis-base~\cite{dynamicvis} & 92.80 & \cellcolor{gray}69.70 & 75.40 & 61.40 & 80.70 & 72.40 & 43.40 & \cellcolor{lightgray}15.90 & 30.10 & 68.50 & 90.80 & 79.60 \\
DynamicVis-large~\cite{dynamicvis} & 92.70 & \cellcolor{darkgray}71.00 & \cellcolor{lightgray}75.80 & 62.10 & 82.00 & 72.80 & \cellcolor{lightgray}45.50 & 15.20 & 42.60 & 69.10 & \cellcolor{lightgray}93.10 & 80.80 \\
\midrule
\textbf{LEVIRDetNet (+0.69, +0.97, +12.69, +2.44)} &
\cellcolor{darkgray}\textbf{98.27} & \textbf{58.29} & \textbf{72.25} &
\cellcolor{darkgray}\textbf{73.55} & \cellcolor{darkgray}\textbf{96.52} & \cellcolor{darkgray}\textbf{88.01} &
\cellcolor{darkgray}\textbf{60.39} & \cellcolor{darkgray}\textbf{26.86} & \cellcolor{darkgray}\textbf{53.41} &
\cellcolor{darkgray}\textbf{76.04} & \textbf{93.08} & \cellcolor{gray}\textbf{85.09} \\
\bottomrule
\end{tabular}}
\end{table*}

\section{Experiments and Analysis}

\subsection{Experimental Setup}
\subsubsection{Experimental Protocol}

We evaluate LEVIRDetNet under a target-training-free cross-benchmark protocol, where LEVIRDetNet is trained on LEVIRDet-159 and directly tested on external benchmarks without target-dataset fine-tuning. This setting is stricter than conventional supervised benchmark evaluation, since a single model must handle unseen imagery, scene distributions, and annotation conventions with one set of weights.

To ensure a valid evaluation, we follow the same split-control rules used in dataset construction. External benchmarks are evaluated with their original images and annotations, and no evaluation image, tile, or parent scene is included in the LEVIRDet-159 training split. For source datasets that also contribute to LEVIRDet-159, we preserve their official partitions and exclude test or validation imagery from training. For example, the NWPU-VHR-10 and HRPlanev2 test images evaluated here are not integrated into LEVIRDet-159 and are assessed using their original benchmark annotations rather than LEVIRDet-adjusted labels.

We organize the comparisons into two groups to avoid ambiguity between target-supervised evaluation and direct-transfer evaluation. LEVIRDetNet is always evaluated under the direct-transfer setting: it is trained only on LEVIRDet-159 and is directly tested on each external benchmark without using the target benchmark training split, target-domain fine-tuning, or target-specific adaptation. In contrast, all supervised baselines are evaluated in the conventional target-supervised setting. For each benchmark, these baselines are trained on the official training split of that benchmark and evaluated using the corresponding official evaluation protocol. We reproduce all baseline results using the released implementations and recommended training settings whenever the code is available. For methods without publicly available code, we report the results from the original papers or official benchmark records and mark them accordingly. This setting intentionally compares a single universal detector without target-domain training against target-supervised detectors trained separately for each benchmark, thereby providing a stringent evaluation of cross-dataset generalization.

We compare the proposed LEVIRDetNet with other state-of-the-art approaches, representative supervised and Transformer-based detectors, including Fast R-CNN\cite{fastrcnn}, Faster R-CNN\cite{FasterR-CNN}, HrNet\cite{hrnet}, Sparse R-CNN\cite{sparse_rcnn}, RTMDet\cite{rtmdet}, YOLO variants\cite{yolo,yolof,yolosiou,yolov12,yolov5,yolox,mha_yolov5}, EfficientDet\cite{efficientdet}, CenterNet\cite{centernet}, Deformable DETR\cite{deformable_detr}, DINO\cite{dino}, FCOS\cite{fcos}, RetinaNet\cite{retinanet}, SSD\cite{ssd}, TridentNet\cite{tridentnet}, DynamicVis\cite{dynamicvis}, DEIM\cite{deim}, and DEIMv2\cite{deimv2} with DINOv3\cite{dinov3}. In addition, we include dataset-specific state-of-the-art methods that are specially designed or extensively reported for their corresponding benchmarks\cite{dssd,fs_ssd,fssd,rfcn,atss,gflv2,vfnet,querydet,czdet,mha_yolov5,dstda,hossein}, enabling a more comprehensive comparison under each dataset setting. We further compare with open-vocabulary or open-set methods, including LAE-DINO\cite{laedino} and SAM3\cite{sam3}, using threshold-sweep evaluation. For these methods, we report the full score-threshold curve from 0 to 1, the best point on each curve, and representative operating points at thresholds 0.3, 0.5, and 0.7.

For supervised baselines trained in our environment, we follow the official paper or codebase recipes and use the longest default schedule available for each method. The same target training data and evaluation protocol are used within each comparison group, with only necessary adaptations to class mappings and input formats. This ensures that the comparison reflects model generalization rather than differences in data leakage, annotation rewriting, or shortened optimization.

\subsubsection{Benchmark Datasets and Evaluation Metrics}

We evaluate LEVIRDetNet on nine external remote sensing benchmarks covering aircraft, vehicles, ships, and general aerial objects, including ADCOS\cite{adcos}, HRPlanev2\cite{hrplane}, SkyFusion-plane\cite{skyfusion}, UCAS-AOD\cite{ucasaod}, CORS-ADD\cite{corsadd}, VHRV\cite{vhrv}, NWPU-VHR-10\cite{NWPU}, CarPK\cite{carpk}, and SkyFusion-ship\cite{skyfusion}. NWPU-VHR-10 and HRPlanev2 are related to sources used in LEVIRDet-159 construction, but their benchmark test images are excluded from LEVIRDet-159 and are evaluated with their original annotations. UCAS-AOD is evaluated by converting oriented boxes to horizontal boxes, while the remaining datasets are not included in the LEVIRDet-159 training set and are used to assess cross-dataset generalization. A dash denotes a metric not reported in the source publication\cite{rfcn,dssd,fssd,atss,gflv2,vfnet,querydet,czdet,mha_yolov5,dstda,hossein,yolov5,yolosiou}, all other baseline results are reproduced by us.

We report standard COCO-style detection metrics, including $\text{mAP}$, $\text{AP}^{50}_{\text{bbox}}$, and $\text{AP}^{75}_{\text{bbox}}$ when available. For datasets with size-specific protocols, we additionally report APs and APm. For threshold-sweep analysis, we evaluate $\text{mAP}$, $\text{AP}^{50}_{\text{bbox}}$, precision, and recall over score thresholds from 0 to 1, which avoids relying on a single manually selected confidence threshold.

Although LEVIRDetNet uses a hierarchy-aware head, all external benchmark comparisons are conducted under a strict single-label protocol. Except for NWPU-VHR-10, each prediction is collapsed to one category before evaluation and matched only to the corresponding benchmark label. We do not use ancestor-descendant compatibility, mixed-depth matching, or hierarchy-specific ignore rules, ensuring direct comparability with standard flat detectors and grounding models.

NWPU-VHR-10 is the only exception and is evaluated with a class-wise protocol because its original annotations are not fully dense across all visible categories. Joint multi-class evaluation could count reasonable detections of unlabeled objects as false positives. Therefore, for each target class, we evaluate only the corresponding category-specific image subset and then average over classes. This protocol is used only for NWPU-VHR-10 and is kept separate from the stricter cross-dataset and threshold-sweep evaluations.

\begin{figure*}
\centering
\includegraphics[width=1\linewidth]{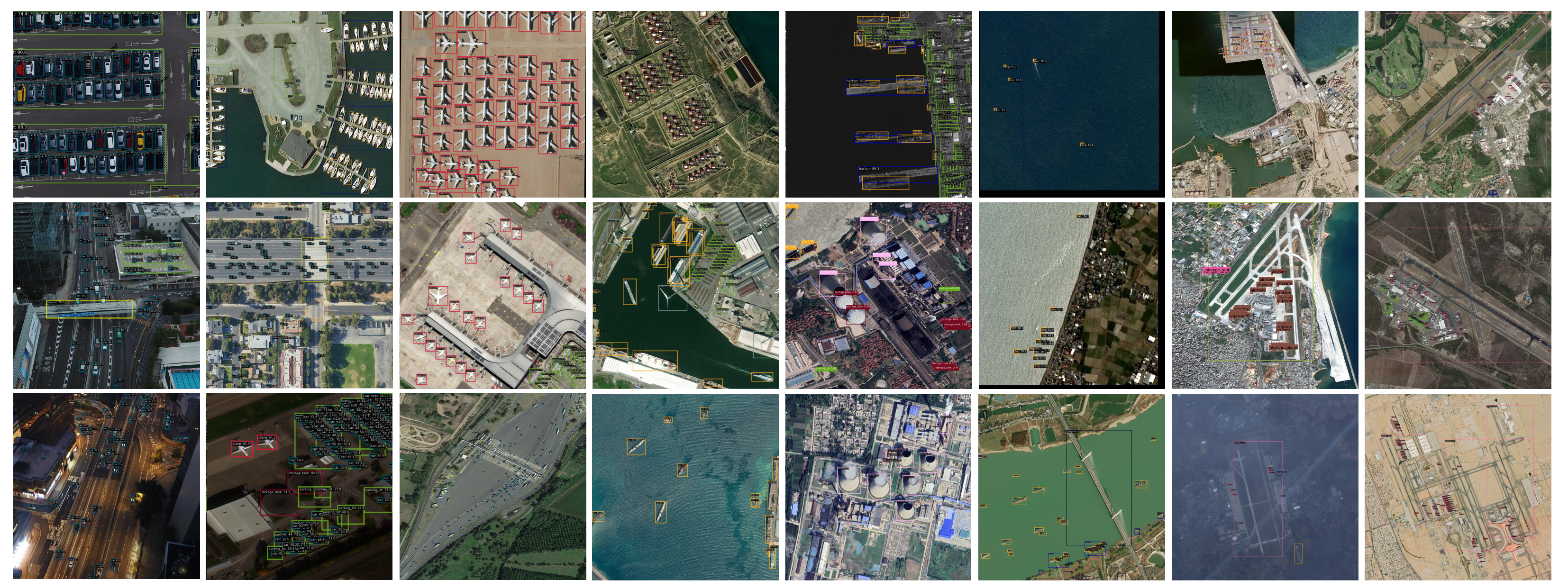}
\vspace{-4ex}
\caption{
Visualization of LEVIRDetNet on unseen external remote sensing images outside LEVIRDet-159. Each image is processed by a single 1024$\times$1024 forward pass without tiling. The results show strong generalization across diverse scenes, categories, resolutions, and imaging conditions without target-domain training.
} 
\label{fig:vislization}
\vspace{-2ex}
\end{figure*}

\begin{table}[!t]
\caption{Comparison with other methods on the CARPK dataset\cite{carpk}.}
\label{tab:comparisons_carpk}
\centering
\resizebox{\linewidth}{!}{
\begin{tabular}{c|c c c}
\toprule
\multicolumn{1}{c|}{\shortstack{Method}} &
$\text{AP}^{50}_{\text{bbox}}$ &
$\text{AP}^{m}_{\text{bbox}}$ &
$\text{AP}^{75}_{\text{bbox}}$ \\
\midrule
\multicolumn{1}{c|}{HrNet\cite{hrnet}} & 80.95 & 52.87 & 61.29 \\
\multicolumn{1}{c|}{Fast R-CNN\cite{fastrcnn}} & 80.07 & 52.78 & 60.76 \\
\multicolumn{1}{c|}{YOLOv3\cite{yolo}} & 85.46 & 43.33 & 60.97 \\
\multicolumn{1}{c|}{YOLOF\cite{yolof}} & 74.16 & 36.78 & 55.05 \\
\multicolumn{1}{c|}{Sparse R-CNN\cite{sparse_rcnn}} & 48.92 & 25.10 & 21.17 \\
\multicolumn{1}{c|}{RTMDet\cite{rtmdet}} & 91.87 & 57.37 & \cellcolor{lightgray}89.62 \\
\multicolumn{1}{c|}{EfficientDet\cite{efficientdet}} & 76.87 & 41.44 & 39.28 \\
\multicolumn{1}{c|}{CenterNet\cite{centernet}} & 81.09 & 54.33 & 64.40 \\
\multicolumn{1}{c|}{Deformable DETR\cite{deformable_detr}} & 77.37 & 44.82 & 47.25 \\
\multicolumn{1}{c|}{DINO\cite{dino}} & 90.14 & 60.16 & 70.81 \\
\multicolumn{1}{c|}{FCOS\cite{fcos}} & 77.98 & 52.05 & 60.99 \\
\multicolumn{1}{c|}{RetinaNet\cite{retinanet}} & 77.72 & 48.27 & 52.90 \\
\multicolumn{1}{c|}{TridentNet\cite{tridentnet}} & 89.83 & 47.48 & 82.26 \\
\multicolumn{1}{c|}{SSD\cite{ssd}} & 90.15 & 50.66 & 50.58 \\
\multicolumn{1}{c|}{Faster R-CNN\cite{FasterR-CNN}} & 80.96 & 53.28 & 63.24 \\
\multicolumn{1}{c|}{YOLOX\cite{yolox}} & 93.86 & 58.87 & 66.60 \\
\multicolumn{1}{c|}{R-FCN\cite{rfcn}} & 86.13 & -- & -- \\
\multicolumn{1}{c|}{DSSD\cite{dssd}} & 87.27 & -- & -- \\
\multicolumn{1}{c|}{FSSD\cite{fssd}} & 87.59 & -- & -- \\
\multicolumn{1}{c|}{FS-SSD\cite{fs_ssd}} & 89.52 & -- & -- \\
\multicolumn{1}{c|}{ATSS\cite{atss}} & 95.94 & -- & -- \\
\multicolumn{1}{c|}{GFLV2\cite{gflv2}} & 94.91 & -- & -- \\
\multicolumn{1}{c|}{VFNet\cite{vfnet}} & 94.97 & -- & -- \\
\multicolumn{1}{c|}{QueryDet\cite{querydet}} & 93.96 & -- & -- \\
\multicolumn{1}{c|}{CZ Det\cite{czdet}} & 92.18 & -- & -- \\
\multicolumn{1}{c|}{MHA-YOLOv5\cite{mha_yolov5}} & \cellcolor{lightgray}97.95 & -- & -- \\
\multicolumn{1}{c|}{DSTDA\cite{dstda}} & \cellcolor{gray}98.20 & -- & -- \\
\multicolumn{1}{c|}{Hossein\cite{hossein}} & 97.20 & -- & -- \\
\multicolumn{1}{c|}{YOLOv12x\cite{yolov12}} & 97.29 & \cellcolor{lightgray}71.35 & 84.41 \\
\multicolumn{1}{c|}{DEIM\cite{deim}} & 94.25 & 67.23 & 80.27 \\
\multicolumn{1}{c|}{DEIMv2 (DINOv3 + CNN)\cite{deimv2}} & 91.06 & 64.37 & 79.13 \\
\multicolumn{1}{c|}{DEIMv2 (DINOv3 + ViT-Base)\cite{deimv2}} & 96.74 & \cellcolor{gray}74.21 & \cellcolor{gray}89.68 \\
\multicolumn{1}{c|}{DynamicVis-base\cite{dynamicvis}} & 78.40 & 48.30 & 51.90 \\
\multicolumn{1}{c|}{DynamicVis-large\cite{dynamicvis}} & 78.70 & 51.50 & 57.90 \\
\midrule
\multicolumn{1}{c|}{\textbf{LEVIRDetNet (+0.59)}} & \cellcolor{darkgray}\textbf{98.79} & \cellcolor{darkgray}\textbf{98.82} & \cellcolor{darkgray}\textbf{90.70} \\
\bottomrule
\end{tabular}}
\end{table}

\subsubsection{Official Validation on LEVIRDet-159}

We also report results on the official LEVIRDet-159 train/validation split to provide a fixed in-dataset benchmark in addition to external transfer evaluation. Table~\ref{tab:levir_det_val} reports the parent-level comparison and the fine-grained LEVIRDetNet results.

\begin{table}[!t]
\caption{
Validation results on the official LEVIRDet-159 validation split. The coarse protocol evaluates the 30 parent categories, while the fine protocol evaluates the complete 159-category taxonomy.
}
\label{tab:levir_det_val}
\centering
\resizebox{\linewidth}{!}{
\begin{tabular}{c|c|c c c c c c}
\toprule
Protocol & Method & mAP & $\text{AP}_{50}$ & $\text{AP}_{75}$ & $\text{AP}_{s}$ & $\text{AP}_{m}$ & $\text{AP}_{l}$ \\
\midrule
Coarse & DEIMv2 + DINOv3 & 63.73 & 84.18 & 70.37 & 27.82 & 55.59 & 77.79 \\
Coarse & LEVIRDetNet & \textbf{66.58} & \textbf{85.97} & \textbf{73.28} & \textbf{32.17} & \textbf{58.22} & \textbf{79.65} \\
Coarse & OpenRSD & 28.15 & 45.01 & 28.82 & 10.23 & 23.86 & 36.52 \\
\midrule
Fine & LEVIRDetNet & 60.13 & 71.82 & 66.98 & 27.87 & 57.92 & 66.29 \\
Fine & OpenRSD & 18.94 & 27.19 & 21.98 & 5.45 & 11.78 & 23.28 \\
\bottomrule
\end{tabular}
}
\end{table}

On the 30-category parent-level protocol, LEVIRDetNet improves over DEIMv2+DINOv3 by 2.85 mAP and raises AP$_s$ from 27.82 to 32.17, showing that its gain also holds when both models are trained and evaluated on the same LEVIRDet-159 split. We report the baseline comparison at the coarse level because forcibly flattening the complete hierarchy changes the semantics of parent categories, as analyzed in Table~\ref{tab:ablation_hierarchy_head}. Training every baseline on the full 159-category taxonomy is costly, so this table uses the strongest DEIMv2+DINOv3 baseline as the full-set reference. The 159-category row is therefore reported as the official fine-grained LEVIRDetNet benchmark. OpenRSD\cite{huang2025openrsd} is a recent open vocabulary detection method. It was not fine-tuned on levirdet-159 and was directly used for testing. 



\subsection{Quantitative Analysis}

\subsubsection{Cross-Benchmark Generalization Results}

LEVIRDetNet trained on LEVIRDet-159 demonstrates strong target-training-free cross-benchmark performance across diverse external benchmarks. LEVIRDetNet achieves the best mAP on ADCOS, HRPlanev2, UCAS-AOD, and VHRV, reaching 83.60, 81.92, 80.40, and 73.55 mAP, respectively, and improving the strongest competing baselines by 4.15, 3.05, 5.02, and 0.97 points. It also generalizes well to vehicle- and ship-centric datasets, obtaining 90.70 AP75 and 98.79 AP50 on CarPK, 60.39 AP50 on SkyFusion-ship, and 98.27 AP50 on SkyFusion-plane, with clear gains over the strongest competing methods on the latter two benchmarks. These results indicate that large-scale fine-grained supervision and scale-aware modeling improve cross-domain detection rather than only fitting the training benchmark.

Under the class-wise NWPU-VHR-10 protocol, LEVIRDetNet achieves 76.04 mAP, the best among all compared methods. Although DEIMv2+DINOv3 obtains slightly higher AP50 and AP75 on this dataset, LEVIRDetNet shows stronger overall precision-recall behavior under the full COCO-style metric. This suggests that cross-benchmark evaluation should rely on mAP rather than AP50 alone, especially when benchmarks differ in annotation density and localization strictness.

All evaluation images are excluded from training. After also excluding NWPU-VHR-10, HRPlanev2, and UCAS-AOD, whose training images contributes to LEVIRDet-159 training split, the gain over the best competing method-level mean is 4.715 points across the remaining six datasets. When grouped by each benchmark's primary metric, LEVIRDetNet improves over the strongest competing method-level mean by 4.35 mAP points across six datasets and 6.34 AP$_{50}$ points across three datasets.

\subsubsection{Comparison With Open-Set Detectors}

\begin{figure*}
\centering
\includegraphics[width=1\linewidth]{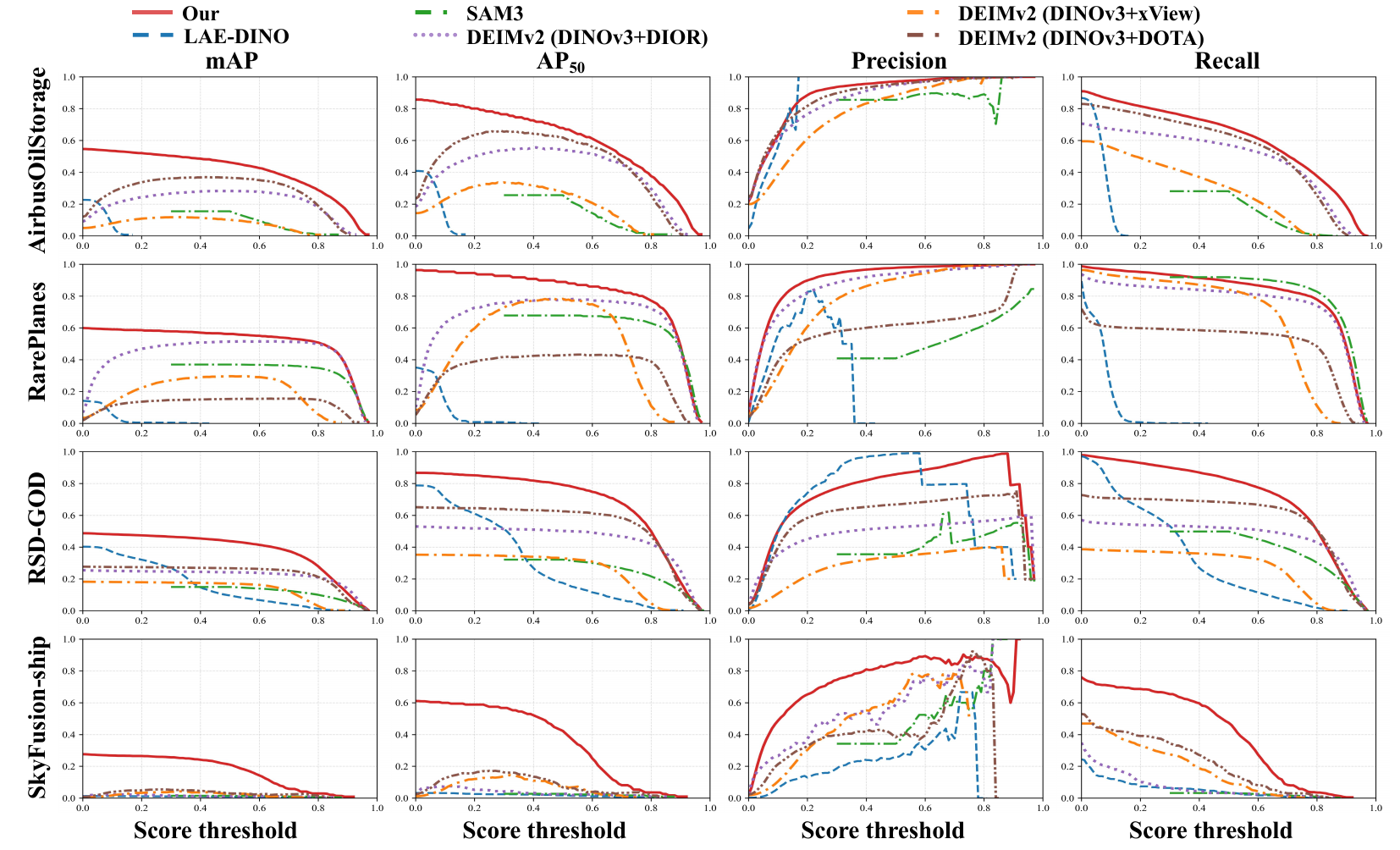}
\vspace{-4ex}
\caption{
A comparison between LEVIRDetNet and open-set/grounding baselines at different score thresholds. Each row reports mAP, AP$_{50}$, precision, and recall on four external benchmarks, illustrating both peak performance and practical-threshold stability.}
\label{fig:score_threshold_metric_curves_4x4_all_methods}
\vspace{-1ex}
\end{figure*}
\begin{table*}[!t]
\caption{
A threshold-sweep peak mAP comparison on four external benchmarks.
}
\label{tab:threshold_sweep_peak_map}
\centering
\resizebox{\linewidth}{!}{
\begin{tabular}{c|c c c c c c}
\toprule
Dataset 
& {LAE-DINO\cite{laedino}} 
& {SAM3\cite{sam3}} 
& {DEIMv2-DIOR\cite{deimv2}} 
& {DEIMv2-xView\cite{deimv2}} 
& {DEIMv2-DOTA\cite{deimv2}} 
& {LEVIRDetNet (Our)} \\
\midrule
AOSD\cite{AOSD}         & 22.66 & 15.48 & 28.26 & 11.82 & 36.97 & \textbf{54.63 (+17.65)}   \\
RarePlanes\cite{rareplanes}   & 14.16 & 36.92 & 51.59 & 29.62 & 15.57 & \textbf{59.93 (+8.34)}  \\
RSD-GOD\cite{gsdgod}      & 40.25 & 14.92 & 25.51 & 18.21 & 27.71& \textbf{48.71 (+8.46)}   \\
SkyFusion-ship\cite{skyfusion}   & 1.34  & 1.48  & 2.34  & 4.27  & 5.52   & \textbf{27.63 (+22.11)}  \\
\bottomrule
\end{tabular}
}
\end{table*}

We further compare LEVIRDetNet with open-set and open-vocabulary methods under score-threshold sweeps. This setting is well aligned with grounding models, since they can be prompted with new category names and applied to unseen datasets without target-domain fine-tuning. The key question is therefore whether their score calibration, localization quality, and small-object behavior are reliable enough for remote sensing detection.

Threshold sweeping is necessary because confidence scores have different meanings across detector families. In closed-set detectors, the score is produced by a fixed classifier, whereas in grounding models it may reflect text-image alignment, prompt compatibility, mask quality, or post-processing. Generic recommended thresholds may suppress most detections in remote sensing scenes, while overly low thresholds can introduce many false positives. We therefore report the full score-threshold curves on AOSD\cite{AOSD}, RarePlanes\cite{rareplanes}, RSD-GOD\cite{gsdgod}, and SkyFusion-ship\cite{skyfusion}, including the best point on each curve as an oracle-threshold upper bound and representative operating points. SAM3 is evaluated using the official detection demo that converts masks to boxes. Since a zero threshold does not produce a meaningful box set in this pipeline, its curve starts from 0.3.

For open-set and grounding baselines, we use a fixed prompt dictionary and duplicate-removal protocol for all threshold values. The final prompt-to-class mapping is: \{plane, airplane\}$\rightarrow$plane; \{tank, oiltank, storagetank\}$\rightarrow$oiltank; \{airbase, airport\}$\rightarrow$airport; ship$\rightarrow$ship; warship$\rightarrow$warship; and helicopter$\rightarrow$helicopter. Duplicate boxes produced by multiple prompts are removed using NMS with an IoU threshold of 0.5, and at most 300 detections are retained per image. The same prompt dictionary, synonym-merging rules, NMS setting, and threshold grid are used for all compared open-set methods.

LEVIRDetNet achieves the best peak mAP on all four threshold-sweep benchmarks, as shown in Fig. \ref{fig:score_threshold_metric_curves_4x4_all_methods}. It reaches 54.63 mAP on AOSD, outperforming the strongest competing curve by 17.65 points; 59.93 mAP on RarePlanes, exceeding the best baseline by 8.34 points; 48.71 mAP on RSD-GOD, improving over LAE-DINO by 8.46 points; and 27.63 mAP on SkyFusion-ship, surpassing the strongest competing curve by 22.11 points. These results show that LEVIRDetNet maintains strong transfer performance across aircraft-centric, general object, and challenging cross-source scenarios.

The curve shapes further reveal different failure modes. Grounding models often preserve more candidates at very low thresholds and can outperform single-dataset closed-set baselines in this range, reflecting their advantage in open-category transfer. However, their performance usually drops rapidly as the threshold increases, indicating weak score calibration for remote sensing objects. In contrast, single-dataset DEIMv2 baselines often show a rise-then-fall pattern. Thresholding removes low-score false positives at first, but later suppresses true positives as well. Their best source dataset also varies across benchmarks, suggesting brittle cross-dataset transfer.

LEVIRDetNet is more stable at practical thresholds. At threshold 0.3, it obtains 50.30, 57.79, 46.50, and 25.65 mAP on AirbusOilStorage (AOSD), RarePlanes, RSD-GOD, and SkyFusion-ship, respectively. At threshold 0.5, it still keeps high precision and useful recall, achieving 97.1\%/68.3\%, 97.8\%/89.3\%, 85.8\%/82.8\%, and 85.1\%/47.4\% precision/recall on the four datasets. This indicates that LEVIRDetNet combines the complementary strengths of the baselines. It preserves cross-dataset recall similar to grounding models while maintaining the precision and score stability of supervised detectors. The comparison also supports our central claim that universal remote sensing detection requires not only large-scale data, but also strict category, geometry, and split unification.

\subsection{Ablation Study}

In this section, we conduct component ablations to validate the effectiveness of our proposed method. Unless otherwise specified, these ablations use a development subset containing 25\% of the full training samples and 50\% of the full validation samples to reduce training cost, so the results should be compared within each ablation table rather than with the full-set validation scores in Table~\ref{tab:levir_det_val}.

\subsubsection{Ablation on GSD Conditioning}

We ablate the effect of GSD-aware conditioning on the validation set. The baseline is a DEIM-style detector without the GSD branch. We then add the predicted visual GSD cue and further introduce a learnable GSD embedding for query-level conditioning. All variants are trained and evaluated under the same settings.

Table~\ref{tab:ablation_gsd_conditioning} shows that GSD conditioning improves mAP from 50.53 to 50.86 and raises AP$_s$ by 1.71 points. The learnable embedding further improves mAP to 51.14 and AP$_s$ to 18.13, indicating that visual scale is most useful for small and medium objects where pixel size and physical object size are weakly aligned. The slight AP$_l$ drop is acceptable because large objects are already well handled by the baseline, while the GSD branch mainly benefits small and medium objects where scale ambiguity is more severe. Overall, the ablation verifies that GSD-aware query conditioning brings consistent gains, particularly for small-object detection and stricter localization quality.




\subsubsection{Ablation on GSD Prediction}

\begin{table}[!t]
\caption{
Ablation study of GSD-aware conditioning on the validation set.
}
\label{tab:ablation_gsd_conditioning}
\centering
\resizebox{\linewidth}{!}{
\begin{tabular}{c|c c c c c c}
\toprule
Method 
& mAP 
& $\text{AP}_{50}$ 
& $\text{AP}_{75}$ 
& $\text{AP}_{s}$ 
& $\text{AP}_{m}$ 
& $\text{AP}_{l}$ \\
\midrule
Baseline 
& 50.53 & 69.63 & 54.85 & 16.25 & 43.12 & \textbf{68.92} \\
Baseline + GSD 
& 50.86 & 69.41 & 55.45 & 17.96 & 43.70 & 67.71 \\
Baseline + GSD + Emb. 
& \textbf{51.14} & \textbf{70.26} & \textbf{55.60} & \textbf{18.13} & \textbf{43.92} & 67.43 \\
\bottomrule
\end{tabular}
}
\end{table}

\begin{table}[!t]
\caption{
Ablation study of different feature designs for GSD estimation on the validation set.
}
\label{tab:ablation_gsd_estimation}
\centering
\resizebox{\linewidth}{!}{
\begin{tabular}{c|c c c c}
\toprule
Method 
& RMSE 
& $\text{Acc}@1$ 
& $\text{Acc}@0.5$ 
& $\text{Acc}@0.2$ \\
\midrule
CNN 
& 0.2568 & 99.06\% & 97.09\% & 79.96\% \\
CNN + FFT 
& 0.2214 & 99.35\% & 96.66\% & 77.88\% \\
CNN + FFT + EDGE 
& \textbf{0.1728} & \textbf{99.51\%} & \textbf{98.08\%} & \textbf{90.19\%} \\
\bottomrule
\end{tabular}
}
\end{table}

Table~\ref{tab:ablation_gsd_estimation} compares visual GSD features. Metrics are computed in the original GSD space (m/pixel); Acc@1 measures the percentage of samples with absolute error below 1 m/pixel. CNN features yield high coarse accuracy but relatively large RMSE. Adding FFT reduces RMSE from 0.2568 to 0.2214 without consistently improving stricter accuracies. Adding edge statistics yields the best results on all metrics, with RMSE of 0.1728 and Acc@0.2 of 90.19\%, suggesting complementary CNN, frequency, and edge-sharpness cues.


\subsubsection{Efficiency and Dynamic Query Allocation Analysis}

We further analyze the effect of dynamic query allocation and the runtime efficiency of LEVIRDetNet. The dynamic query module is not designed as a pure acceleration component. Its main purpose is to adapt the query capacity to image density. Using a small fixed query budget may miss objects in dense scenes, while using a large fixed budget for all images allocates many queries to background regions in sparse scenes and makes optimization less stable. In DETR-style detectors, one-to-one matching is used during training, whereas inference directly decodes object queries without Hungarian matching or back-propagation \cite{detr}. Therefore, the number of queries should be interpreted together with empirical runtime rather than as a standalone inference-cost indicator.

\begin{table}[!t]
\caption{
Ablation study of dynamic query allocation on the validation set.
}
\label{tab:ablation_dynamic_query}
\centering
\resizebox{\linewidth}{!}{
\begin{tabular}{c|c c c c c c}
\toprule
Method 
& mAP 
& $\text{AP}_{50}$ 
& $\text{AP}_{75}$ 
& $\text{AP}_{s}$ 
& $\text{AP}_{m}$ 
& $\text{AP}_{l}$ \\
\midrule
w/o DynQ 
& 50.06 & 69.19 & 54.50 & 18.37 & \textbf{43.49} & 66.35 \\
w/ DynQ 
& \textbf{50.66} & \textbf{69.81} & \textbf{55.31} & \textbf{18.46} & 43.07 & \textbf{67.09} \\
\bottomrule
\end{tabular}}
\end{table}

\begin{table}[!t]
\caption{
Runtime and performance comparison. FPS and latency are measured on a single RTX 4090 with batch size 1. $\text{mAP}_{\text{close}}$ denotes the average primary AP over the nine closed-set external benchmarks, and $\text{mAP}_{\text{open}}$ denotes the average peak mAP on the four threshold-sweep benchmarks.
}
\label{tab:efficiency_analysis}
\centering
\resizebox{\linewidth}{!}{
\begin{tabular}{c|c c c c c}
\toprule
Method 
& FPS (imgs/s)
& Latency (ms) 
& FLOPs (G)
& $\text{mAP}_{\text{close}}$ 
& $\text{mAP}_{\text{open}}$ \\
\midrule
LAE-DINO 
& 9.00 & 111.10 & 300 & -- & 19.60 \\
SAM3 
& 3.86 & 258.87 & 5036.94 & -- & 17.20 \\
YOLOv12x 
& \textbf{96.30} & \textbf{10.38} & 184.6 & 75.17 & -- \\
DEIMv2 
& 18.55 & 53.91 & \textbf{174} & 75.54 & 26.93 \\
LEVIRDetNet 
& 15.20 & 65.78 & 216.0 & \textbf{80.56} & \textbf{47.73} \\
\bottomrule
\end{tabular}}
\end{table}

As shown in Table~\ref{tab:ablation_dynamic_query}, the dynamic-query variant obtains slightly higher overall mAP, $\text{AP}_{s}$ and $\text{AP}_{l}$. This result indicates that dynamic query allocation should not be viewed as an isolated accuracy booster. Instead, it provides an adaptive query-budget mechanism for images with different object densities, avoiding the need to use one fixed query number for both sparse and crowded scenes.

Table~\ref{tab:efficiency_analysis} shows that LEVIRDetNet remains practically efficient despite using a high maximum query budget. It runs at 15.20 FPS on a single RTX 4090 with batch size 1. Compared with DEIMv2, LEVIRDetNet introduces moderate additional cost due to GSD conditioning, dynamic query allocation, and hierarchy-aware classification, but improves the average closed-set external-benchmark AP from 75.54 to 80.56 and the average open-set threshold-sweep peak mAP from 26.93 to 47.73. Compared with LAE-DINO and SAM3, LEVIRDetNet is substantially faster while achieving much stronger open-benchmark performance. These results suggest that the proposed scale-hierarchy-aware design improves transfer performance without making inference impractical.




\subsubsection{Ablation on Hierarchy-Aware Head}

We compare the hierarchy-aware head with a flat head that treats parent, intermediate, and leaf nodes as independent categories while keeping the detector and training data unchanged.

\begin{table}[!t]
\caption{
Ablation study of the hierarchy-aware head. \(L_1\), \(L_2\), and \(L_3\) denote parent, intermediate, and fine-grained levels, respectively.
}
\label{tab:ablation_hierarchy_head}
\centering
\resizebox{\linewidth}{!}{
\begin{tabular}{c|c|c c c|c c c}
\toprule
\multirow{2}{*}{Branch} & \multirow{2}{*}{Level}
& \multicolumn{3}{c|}{Hier. head}
& \multicolumn{3}{c}{Flat head} \\
\cmidrule(lr){3-5}\cmidrule(lr){6-8}
& & mAP & $\text{AP}_{50}$ & $\text{AP}_{75}$ 
& mAP & $\text{AP}_{50}$ & $\text{AP}_{75}$ \\
\midrule
Plane & \(L_1\) & \textbf{83.10} & \textbf{97.50} & \textbf{91.70} & 75.50 & 93.60 & 83.80 \\
Plane & \(L_2\) & \textbf{72.03} & \textbf{82.13} & \textbf{80.00} & 52.13 & 60.17 & 58.50 \\
Plane & \(L_3\) & 65.85 & 70.84 & 69.89 & \textbf{80.91} & \textbf{89.01} & \textbf{87.15} \\
Ship  & \(L_1\) & \textbf{70.10} & \textbf{92.40} & \textbf{78.50} & 58.20 & 71.80 & 63.70 \\
Ship  & \(L_2\) & \textbf{63.85} & \textbf{83.35} & \textbf{75.35} & 57.30 & 74.95 & 66.55 \\
Ship  & \(L_3\) & \textbf{39.85} & \textbf{51.26} & \textbf{46.66} & 37.69 & 48.66 & 43.23 \\
\bottomrule
\end{tabular}
}
\end{table}

Table~\ref{tab:ablation_hierarchy_head} shows the cost of flattening. A flat head changes a parent node from an inclusive family label into a residual class excluding its children, so parent and child supervision compete. The hierarchy-aware head improves aircraft \(L_1/L_2\) mAP by 7.60/19.90 points and ship \(L_1/L_2\) mAP by 11.90/6.55 points. Although the flat head increases aircraft \(L_3\) accuracy, it does so by sacrificing higher-level semantics, and it does not improve fine-grained ship detection. This supports using hierarchy-aware classification for mixed-granularity labels.



\subsection{Discussion and Limitations}

The results indicate that LEVIRDet's gains come from coupling data unification with model design. Scale alone is not sufficient. Heterogeneous annotations must be completed, corrected, and placed under one tight-HBB and hierarchical protocol. On this supervision, LEVIRDetNet benefits from modeling visual scale and mixed-granularity categories, matching the fact that an object's reliable semantic depth depends on image evidence. The direct-transfer results further suggest that LEVIRDet-159 can serve as a training foundation for universal remote sensing object detection.

LEVIRDet still has limitations. The GSD predictor provides a detection-useful visual scale cue, not a replacement for sensor calibration. Fine-grained fallback under severe visual degradation still depends on annotation rules and human review. The current protocol uses horizontal boxes and does not directly address oriented detection. The deepest taxonomy focuses on aircraft, vehicles, and ships. Expanding reliable fine-grained branches for other geospatial objects remains future work.

\section{Conclusion}

This paper presents LEVIRDet, a unified data-and-model framework for universal remote sensing object detection built around LEVIRDet-159, a 159-category million-scale dataset, and LEVIRDetNet, a scale-hierarchy-aware detection foundation model. The LEVIRDet-159 dataset provides 159-category million-scale annotations under a unified tight-HBB protocol, with large-scale instance completion, geometry correction, fine-grained relabeling, and mixed-depth hierarchical labels. Built on this supervision, LEVIRDetNet introduces online visual GSD conditioning, dynamic query allocation, and hierarchy-aware classification to adapt query-based detection to cross-source remote sensing scenes. Comprehensive experiments on nine external benchmarks demonstrate state-of-the-art target-training-free cross-benchmark performance without target-domain fine-tuning, while threshold-sweep comparisons show more stable precision-recall behavior than open-set and grounding models. These results highlight LEVIRDet's central message. High-quality annotation unification and scale-hierarchy-aware model design are complementary foundations for practical universal remote sensing detection.

\ifCLASSOPTIONcaptionsoff
  \newpage
\fi

\bibliographystyle{IEEEtran}

\bibliography{references.bib}

@article{Objectdetectionin20years,
  title={Object detection in 20 years: A survey},
  author={Zou, Zhengxia and Chen, Keyan and Shi, Zhenwei and Guo, Yuhong and Ye, Jieping},
  journal={Proceedings of the IEEE},
  volume={111},
  number={3},
  pages={257--276},
  year={2023},
  publisher={IEEE}
}

@article{hossein,
  title={Fine-tuned YOLOv5 for real-time vehicle detection in UAV imagery: Architectural improvements and performance boost},
  author={Hamzenejadi, Mohammad Hossein and Mohseni, Hadis},
  journal={Expert Systems with Applications},
  volume={231},
  pages={120845},
  year={2023},
  publisher={Elsevier}
}

@article{rssurvey,
  author = "Yang, Qinzhe and Liu, Chenyang and Xu, Jia and Shi, Zhenwei and Zou, Zhengxia",
  title = "State Space Models Meet Remote Sensing: A Survey",
  journal = "SCIENCE CHINA Information Sciences",
  year = "2026",
  doi = "https://doi.org/10.1007/s11432-025-4780-1"
}

@ARTICLE{rs4d,
  author={Yang, Qinzhe and Chen, Keyan and Xu, Jia and Shi, Zhenwei and Zou, Zhengxia},
  journal={IEEE Transactions on Geoscience and Remote Sensing}, 
  title={Efficient Remote Sensing Instance Segmentation with Linear-Time State Space Distilled Visual Foundation Models}, 
  year={2026},
  volume={},
  number={},
  pages={1-1},
  doi={10.1109/TGRS.2026.3696104}}

@ARTICLE{OBBTPAMI,
  author={Xu, Yongchao and Fu, Mingtao and Wang, Qimeng and Wang, Yukang and Chen, Kai and Xia, Gui-Song and Bai, Xiang},
  journal={IEEE Transactions on Pattern Analysis and Machine Intelligence}, 
  title={Gliding Vertex on the Horizontal Bounding Box for Multi-Oriented Object Detection}, 
  year={2021},
  volume={43},
  number={4},
  pages={1452-1459},
  doi={10.1109/TPAMI.2020.2974745}}

@ARTICLE{SMALLTPAMI,
  author={Yang, Xue and Yan, Junchi and Liao, Wenlong and Yang, Xiaokang and Tang, Jin and He, Tao},
  journal={IEEE Transactions on Pattern Analysis and Machine Intelligence}, 
  title={SCRDet++: Detecting Small, Cluttered and Rotated Objects via Instance-Level Feature Denoising and Rotation Loss Smoothing}, 
  year={2023},
  volume={45},
  number={2},
  pages={2384-2399},
  doi={10.1109/TPAMI.2022.3166956}}

@ARTICLE{OBB2TPAMI,
  author={Nie, Guangtao and Huang, Hua},
  journal={IEEE Transactions on Pattern Analysis and Machine Intelligence}, 
  title={Multi-Oriented Object Detection in Aerial Images With Double Horizontal Rectangles}, 
  year={2023},
  volume={45},
  number={4},
  pages={4932-4944},
  doi={10.1109/TPAMI.2022.3191753}}

@ARTICLE{GLHTPAMI,
  author={Li, Yansheng and Luo, Junwei and Zhang, Yongjun and Tan, Yihua and Yu, Jin-Gang and Bai, Song},
  journal={IEEE Transactions on Pattern Analysis and Machine Intelligence}, 
  title={Learning to Holistically Detect Bridges From Large-Size VHR Remote Sensing Imagery}, 
  year={2024},
  volume={46},
  number={12},
  pages={11507-11523},
  doi={10.1109/TPAMI.2024.3393024}}

@ARTICLE{SmallObjectTPAMI,
  author={Cheng, Gong and Yuan, Xiang and Yao, Xiwen and Yan, Kebing and Zeng, Qinghua and Xie, Xingxing and Han, Junwei},
  journal={IEEE Transactions on Pattern Analysis and Machine Intelligence}, 
  title={Towards Large-Scale Small Object Detection: Survey and Benchmarks}, 
  year={2023},
  volume={45},
  number={11},
  pages={13467-13488},
  doi={10.1109/TPAMI.2023.3290594}}

@ARTICLE{ObjectDetectioninAerialImagesTPAMI,
  author={Ding, Jian and Xue, Nan and Xia, Gui-Song and Bai, Xiang and Yang, Wen and Yang, Michael Ying and Belongie, Serge and Luo, Jiebo and Datcu, Mihai and Pelillo, Marcello and Zhang, Liangpei},
  journal={IEEE Transactions on Pattern Analysis and Machine Intelligence}, 
  title={Object Detection in Aerial Images: A Large-Scale Benchmark and Challenges}, 
  year={2022},
  volume={44},
  number={11},
  pages={7778-7796},
  doi={10.1109/TPAMI.2021.3117983}}

@article{MTARSI,
  title={A benchmark data set for aircraft type recognition from remote sensing images},
  author={Wu, ZhiZe and Wan, ShouHong and Wang, XiaoFeng and Tan, Ming and Zou, Le and Li, XinLu and Chen, Yan},
  journal={Applied Soft Computing},
  volume={89},
  pages={106132},
  year={2020},
  publisher={Elsevier}
}

@article{NWD,
    title={Detecting Tiny Objects in Aerial Images: A Normalized Wasserstein Distance and A New Benchmark},
    author={Xu, Chang and Wang, Jinwang and and Yang, Wen and Yu, Huai and Yu, Lei and Xia, Gui-Song},
    journal={ISPRS Journal of Photogrammetry and Remote Sensing},
    year={2022}
}

@INPROCEEDINGS{fpn,
  author={Lin, Tsung-Yi and Dollár, Piotr and Girshick, Ross and He, Kaiming and Hariharan, Bharath and Belongie, Serge},
  booktitle={2017 IEEE Conference on Computer Vision and Pattern Recognition (CVPR)}, 
  title={Feature Pyramid Networks for Object Detection}, 
  year={2017},
  volume={},
  number={},
  pages={936-944},
  doi={10.1109/CVPR.2017.106}}

@article{DIOR,
  title={Object detection in optical remote sensing images: A survey and a new benchmark},
  author={Li, Ke and Wan, Gang and Cheng, Gong and Meng, Liqiu and Han, Junwei},
  journal={ISPRS journal of photogrammetry and remote sensing},
  volume={159},
  pages={296--307},
  year={2020},
  publisher={Elsevier}
}

@article{Ningbo,
  title={A new spatial-oriented object detection framework for remote sensing images},
  author={Yu, Dawen and Ji, Shunping},
  journal={IEEE Transactions on Geoscience and Remote Sensing},
  volume={60},
  pages={1--16},
  year={2021},
  publisher={IEEE}
}

@article{vhrships,
  title={VHRShips: An extensive benchmark dataset for scalable deep learning-based ship detection applications},
  author={K{\i}z{\i}lkaya, Serdar and Alganci, Ugur and Sertel, Elif},
  journal={ISPRS International Journal of Geo-Information},
  volume={11},
  number={8},
  pages={445},
  year={2022},
  publisher={MDPI}
}

@article{Vaihingen,
  title={Use of the stair vision library within the ISPRS 2D semantic labeling benchmark (Vaihingen)},
  author={Markus Gerke, ITC},
  journal={Use of the stair vision library within the isprs 2d semantic labeling benchmark (vaihingen)},
  year={2014}
}

@article{mar20,
  title={MAR20: A benchmark for military aircraft recognition in remote sensing images},
  author={Wenqi, YU and Gong, Cheng and Meijun, Wang and Yanqing, Yao and Xingxing, Xie and Xiwen, Yao and Junwei, Han},
  journal={National Remote Sensing Bulletin},
  volume={27},
  number={12},
  pages={2688--2696},
  year={2024},
  publisher={National Remote Sensing Bulletin}
}

@inproceedings{FMOW,
  title={Functional map of the world},
  author={Christie, Gordon and Fendley, Neil and Wilson, James and Mukherjee, Ryan},
  booktitle={Proceedings of the IEEE Conference on Computer Vision and Pattern Recognition},
  pages={6172--6180},
  year={2018}
}

@inproceedings{OIRDS,
  title={Overhead imagery research data set—an annotated data library \& tools to aid in the development of computer vision algorithms},
  author={Tanner, Franklin and Colder, Brian and Pullen, Craig and Heagy, David and Eppolito, Michael and Carlan, Veronica and Oertel, Carsten and Sallee, Phil},
  booktitle={2009 IEEE Applied Imagery Pattern Recognition Workshop (AIPR 2009)},
  pages={1--8},
  year={2009},
  organization={IEEE}
}

@article{MASATI,
  title={Automatic ship classification from optical aerial images with convolutional neural networks},
  author={Gallego, Antonio-Javier and Pertusa, Antonio and Gil, Pablo},
  journal={Remote Sensing},
  volume={10},
  number={4},
  pages={511},
  year={2018},
  publisher={MDPI}
}

@article{DLR3kDLR,
  title={Fast multiclass vehicle detection on aerial images},
  author={Liu, Kang and Mattyus, Gellert},
  journal={IEEE Geoscience and Remote Sensing Letters},
  volume={12},
  number={9},
  pages={1938--1942},
  year={2015},
  publisher={IEEE}
}

@article{ITCVD,
  title={Vehicle detection in aerial images},
  author={Yang, Michael Ying and Liao, Wentong and Li, Xinbo and Cao, Yanpeng and Rosenhahn, Bodo},
  journal={Photogrammetric Engineering \& Remote Sensing},
  volume={85},
  number={4},
  pages={297--304},
  year={2019},
  publisher={American Society for Photogrammetry and Remote Sensing}
}

@article{vedai,
  title={Vehicle detection in aerial imagery: A small target detection benchmark},
  author={Razakarivony, Sebastien and Jurie, Frederic},
  journal={Journal of Visual Communication and Image Representation},
  volume={34},
  pages={187--203},
  year={2016},
  publisher={Elsevier}
}

@inproceedings{tas,
  title={Learning spatial context: Using stuff to find things},
  author={Heitz, Geremy and Koller, Daphne},
  booktitle={European conference on computer vision},
  pages={30--43},
  year={2008},
  organization={Springer}
}

@article{hrssd,
  title={Hierarchical and robust convolutional neural network for very high-resolution remote sensing object detection},
  author={Zhang, Yuanlin and Yuan, Yuan and Feng, Yachuang and Lu, Xiaoqiang},
  journal={IEEE Transactions on Geoscience and Remote Sensing},
  volume={57},
  number={8},
  pages={5535--5548},
  year={2019},
  publisher={IEEE}
}

@article{rsod,
  title={Accurate object localization in remote sensing images based on convolutional neural networks},
  author={Long, Yang and Gong, Yiping and Xiao, Zhifeng and Liu, Qing},
  journal={IEEE Transactions on Geoscience and Remote Sensing},
  volume={55},
  number={5},
  pages={2486--2498},
  year={2017},
  publisher={IEEE}
}

@article{levirship,
  title={A degraded reconstruction enhancement-based method for tiny ship detection in remote sensing images with a new large-scale dataset},
  author={Chen, Jianqi and Chen, Keyan and Chen, Hao and Zou, Zhengxia and Shi, Zhenwei},
  journal={IEEE Transactions on Geoscience and Remote Sensing},
  volume={60},
  pages={1--14},
  year={2022},
  publisher={IEEE}
}

@article{shiprsimagenet,
  title={ShipRSImageNet: A large-scale fine-grained dataset for ship detection in high-resolution optical remote sensing images},
  author={Zhang, Zhengning and Zhang, Lin and Wang, Yue and Feng, Pengming and He, Ran},
  journal={IEEE Journal of Selected Topics in Applied Earth Observations and Remote Sensing},
  volume={14},
  pages={8458--8472},
  year={2021},
  publisher={IEEE}
}

@article{fair1m,
  title={FAIR1M: A benchmark dataset for fine-grained object recognition in high-resolution remote sensing imagery},
  author={Sun, Xian and Wang, Peijin and Yan, Zhiyuan and Xu, Feng and Wang, Ruiping and Diao, Wenhui and Chen, Jin and Li, Jihao and Feng, Yingchao and Xu, Tao and others},
  journal={ISPRS Journal of Photogrammetry and Remote Sensing},
  volume={184},
  pages={116--130},
  year={2022},
  publisher={Elsevier}
}

@inproceedings{dota,
  title={DOTA: A large-scale dataset for object detection in aerial images},
  author={Xia, Gui-Song and Bai, Xiang and Ding, Jian and Zhu, Zhen and Belongie, Serge and Luo, Jiebo and Datcu, Mihai and Pelillo, Marcello and Zhang, Liangpei},
  booktitle={Proceedings of the IEEE conference on computer vision and pattern recognition},
  pages={3974--3983},
  year={2018}
}

@inproceedings{HRSC2016,
  title={A high resolution optical satellite image dataset for ship recognition and some new baselines},
  author={Liu, Zikun and Yuan, Liu and Weng, Lubin and Yang, Yiping},
  booktitle={International conference on pattern recognition applications and methods},
  volume={2},
  pages={324--331},
  year={2017},
  organization={SciTePress}
}

@inproceedings{OWLVIT,
  title={Simple open-vocabulary object detection},
  author={Minderer, Matthias and Gritsenko, Alexey and Stone, Austin and Neumann, Maxim and Weissenborn, Dirk and Dosovitskiy, Alexey and Mahendran, Aravindh and Arnab, Anurag and Dehghani, Mostafa and Shen, Zhuoran and others},
  booktitle={European conference on computer vision},
  pages={728--755},
  year={2022},
  organization={Springer}
}

@inproceedings{huang2025openrsd,
  title={Openrsd: Towards open-prompts for object detection in remote sensing images},
  author={Huang, Ziyue and Feng, Yongchao and Liu, Ziqi and Yang, Shuai and Liu, Qingjie and Wang, Yunhong},
  booktitle={2025 IEEE/CVF International Conference on Computer Vision (ICCV)},
  pages={8384--8394},
  year={2025},
  organization={IEEE}
}

@inproceedings{perez2018film,
  title={Film: Visual reasoning with a general conditioning layer},
  author={Perez, Ethan and Strub, Florian and De Vries, Harm and Dumoulin, Vincent and Courville, Aaron},
  booktitle={Proceedings of the AAAI conference on artificial intelligence},
  volume={32},
  number={1},
  year={2018}
}

@inproceedings{groundingdino,
  title={Grounding dino: Marrying dino with grounded pre-training for open-set object detection},
  author={Liu, Shilong and Zeng, Zhaoyang and Ren, Tianhe and Li, Feng and Zhang, Hao and Yang, Jie and Jiang, Qing and Li, Chunyuan and Yang, Jianwei and Su, Hang and others},
  booktitle={European conference on computer vision},
  pages={38--55},
  year={2024},
  organization={Springer}
}

@article{dinov3,
  title={Dinov3},
  author={Sim{\'e}oni, Oriane and Vo, Huy V and Seitzer, Maximilian and Baldassarre, Federico and Oquab, Maxime and Jose, Cijo and Khalidov, Vasil and Szafraniec, Marc and Yi, Seungeun and Ramamonjisoa, Micha{\"e}l and others},
  journal={arXiv preprint arXiv:2508.10104},
  year={2025}
}

@inproceedings{laedino,
  title={Locate anything on earth: Advancing open-vocabulary object detection for remote sensing community},
  author={Pan, Jiancheng and Liu, Yanxing and Fu, Yuqian and Ma, Muyuan and Li, Jiahao and Paudel, Danda Pani and Van Gool, Luc and Huang, Xiaomeng},
  booktitle={Proceedings of the AAAI Conference on Artificial Intelligence},
  volume={39},
  number={6},
  pages={6281--6289},
  year={2025}
}

@article{sam3,
  title={Sam 3: Segment anything with concepts},
  author={Carion, Nicolas and Gustafson, Laura and Hu, Yuan-Ting and Debnath, Shoubhik and Hu, Ronghang and Suris, Didac and Ryali, Chaitanya and Alwala, Kalyan Vasudev and Khedr, Haitham and Huang, Andrew and others},
  journal={arXiv preprint arXiv:2511.16719},
  year={2025}
}

@article{gsdgod,
  title={A single shot framework with multi-scale feature fusion for geospatial object detection},
  author={Zhuang, Shuo and Wang, Ping and Jiang, Boran and Wang, Gang and Wang, Cong},
  journal={Remote Sensing},
  volume={11},
  number={5},
  pages={594},
  year={2019},
  publisher={MDPI}
}

@inproceedings{rareplanes,
  title={Rareplanes: Synthetic data takes flight},
  author={Shermeyer, Jacob and Hossler, Thomas and Van Etten, Adam and Hogan, Daniel and Lewis, Ryan and Kim, Daeil},
  booktitle={Proceedings of the IEEE/CVF Winter Conference on Applications of Computer Vision},
  pages={207--217},
  year={2021}
}

@article{AOSD,
  title={Airbus Oil Storage Detection},
  author={Airbus},
  journal={Kaggle},
  year={2021}
}

@inproceedings{bridge,
  title={A tool for bridge detection in major infrastructure works using satellite images},
  author={Nogueira, Keiller and Cesar, Caio and Gama, Pedro HT and Machado, Gabriel LS and dos Santos, Jefersson A},
  booktitle={2019 XV Workshop de Vis{\~a}o Computacional (WVC)},
  pages={72--77},
  year={2019},
  organization={IEEE}
}

@article{AirbusShip,
  title={Airbus Ship Detection Challenge},
  author={Airbus},
  journal={Kaggle},
  year={2018}
}

@article{AirbusAircraft,
  title={Airbus Aircraft Detection Challenge},
  author={Airbus},
  journal={Kaggle},
  year={2021}
}

@article{dstda,
  title={Dense small target detection algorithm for UAV aerial imagery},
  author={Lu, Sheng and Guo, Yangming and Long, Jiang and Liu, Zun and Wang, Zhuqing and Li, Ying},
  journal={Image and Vision Computing},
  volume={156},
  pages={105485},
  year={2025},
  publisher={Elsevier}
}

@article{mha_yolov5,
  title={Small object detection in unmanned aerial vehicle images using multi-scale hybrid attention},
  author={Song, Gang and Du, Hongwei and Zhang, Xinyue and Bao, Fangxun and Zhang, Yunfeng},
  journal={Engineering Applications of Artificial Intelligence},
  volume={128},
  pages={107455},
  year={2024},
  publisher={Elsevier}
}

@article{rfcn,
  title={R-fcn: Object detection via region-based fully convolutional networks},
  author={Dai, Jifeng and Li, Yi and He, Kaiming and Sun, Jian},
  journal={Advances in neural information processing systems},
  volume={29},
  year={2016}
}

@article{dssd,
  title={Dssd: Deconvolutional single shot detector},
  author={Fu, Cheng-Yang and Liu, Wei and Ranga, Ananth and Tyagi, Ambrish and Berg, Alexander C},
  journal={arXiv preprint arXiv:1701.06659},
  year={2017}
}

@article{fssd,
  title={FSSD: feature fusion single shot multibox detector},
  author={Li, Zuoxin and Yang, Lu and Zhou, Fuqiang},
  journal={arXiv preprint arXiv:1712.00960},
  year={2017}
}

@article{fs_ssd,
  title={Small object detection in unmanned aerial vehicle images using feature fusion and scaling-based single shot detector with spatial context analysis},
  author={Liang, Xi and Zhang, Jing and Zhuo, Li and Li, Yuzhao and Tian, Qi},
  journal={IEEE Transactions on Circuits and Systems for Video Technology},
  volume={30},
  number={6},
  pages={1758--1770},
  year={2019},
  publisher={IEEE}
}

@inproceedings{gflv2,
  title={Generalized focal loss v2: Learning reliable localization quality estimation for dense object detection},
  author={Li, Xiang and Wang, Wenhai and Hu, Xiaolin and Li, Jun and Tang, Jinhui and Yang, Jian},
  booktitle={Proceedings of the IEEE/CVF conference on computer vision and pattern recognition},
  pages={11632--11641},
  year={2021}
}

@inproceedings{atss,
  title={Bridging the gap between anchor-based and anchor-free detection via adaptive training sample selection},
  author={Zhang, Shifeng and Chi, Cheng and Yao, Yongqiang and Lei, Zhen and Li, Stan Z},
  booktitle={Proceedings of the IEEE/CVF conference on computer vision and pattern recognition},
  pages={9759--9768},
  year={2020}
}

@inproceedings{vfnet,
  title={Varifocalnet: An iou-aware dense object detector},
  author={Zhang, Haoyang and Wang, Ying and Dayoub, Feras and Sunderhauf, Niko},
  booktitle={Proceedings of the IEEE/CVF conference on computer vision and pattern recognition},
  pages={8514--8523},
  year={2021}
}

@inproceedings{querydet,
  title={QueryDet: Cascaded sparse query for accelerating high-resolution small object detection},
  author={Yang, Chenhongyi and Huang, Zehao and Wang, Naiyan},
  booktitle={Proceedings of the IEEE/CVF Conference on computer vision and pattern recognition},
  pages={13668--13677},
  year={2022}
}

@inproceedings{czdet,
  title={Cascaded zoom-in detector for high resolution aerial images},
  author={Meethal, Akhil and Granger, Eric and Pedersoli, Marco},
  booktitle={Proceedings of the IEEE/CVF conference on computer vision and pattern recognition},
  pages={2046--2055},
  year={2023}
}

@article{yolov5,
  title={What is YOLOv5: A deep look into the internal features of the popular object detector},
  author={Khanam, Rahima and Hussain, Muhammad},
  journal={arXiv preprint arXiv:2407.20892},
  year={2024}
}

@inproceedings{yolosiou,
  title={You only look once based-c2fghost using efficient siou loss function for airplane detection},
  author={Do, Manh-Tuan and Ha, Manh-Hung and Nguyen, Duc-Chinh and Chen, Oscal Tzyh-Chiang},
  booktitle={2024 9th International Conference on Frontiers of Signal Processing (ICFSP)},
  pages={1--5},
  year={2024},
  organization={IEEE}
}

@ARTICLE{sparse_rcnn,
  author={Sun, Peize and Zhang, Rufeng and Jiang, Yi and Kong, Tao and Xu, Chenfeng and Zhan, Wei and Tomizuka, Masayoshi and Yuan, Zehuan and Luo, Ping},
  journal={IEEE Transactions on Pattern Analysis and Machine Intelligence}, 
  title={Sparse R-CNN: An End-to-End Framework for Object Detection}, 
  year={2023},
  volume={45},
  number={12},
  pages={15650-15664},
  doi={10.1109/TPAMI.2023.3292030}}

@inproceedings{efficientdet,
  title={Efficientdet: Scalable and efficient object detection},
  author={Tan, Mingxing and Pang, Ruoming and Le, Quoc V},
  booktitle={Proceedings of the IEEE/CVF conference on computer vision and pattern recognition},
  pages={10781--10790},
  year={2020}
}

@article{dynamicvis,
  title={Dynamicvis: An efficient and general visual foundation model for remote sensing image understanding},
  author={Chen, Keyan and Liu, Chenyang and Chen, Bowen and Li, Wenyuan and Zou, Zhengxia and Shi, Zhenwei},
  journal={arXiv preprint arXiv:2503.16426},
  year={2025}
}

@article{deimv2,
  title={Real-time object detection meets DINOv3},
  author={Huang, Shihua and Hou, Yongjie and Liu, Longfei and Yu, Xuanlong and Shen, Xi},
  journal={arXiv preprint arXiv:2509.20787},
  year={2025}
}

@article{yolov12,
  title={Yolov12: Attention-centric real-time object detectors},
  author={Tian, Yunjie and Ye, Qixiang and Doermann, David},
  journal={Advances in neural information processing systems},
  volume={38},
  pages={78433--78457},
  year={2026}
}

@inproceedings{tridentnet,
  title={Scale-aware trident networks for object detection},
  author={Li, Yanghao and Chen, Yuntao and Wang, Naiyan and Zhang, Zhaoxiang},
  booktitle={Proceedings of the IEEE/CVF international conference on computer vision},
  pages={6054--6063},
  year={2019}
}

@article{rtmdet,
  title={Rtmdet: An empirical study of designing real-time object detectors},
  author={Lyu, Chengqi and Zhang, Wenwei and Huang, Haian and Zhou, Yue and Wang, Yudong and Liu, Yanyi and Zhang, Shilong and Chen, Kai},
  journal={arXiv preprint arXiv:2212.07784},
  year={2022}
}

@inproceedings{yolof,
  title={You only look one-level feature},
  author={Chen, Qiang and Wang, Yingming and Yang, Tong and Zhang, Xiangyu and Cheng, Jian and Sun, Jian},
  booktitle={Proceedings of the IEEE/CVF conference on computer vision and pattern recognition},
  pages={13039--13048},
  year={2021}
}

@article{yolox,
  title={Yolox: Exceeding yolo series in 2021},
  author={Ge, Zheng and Liu, Songtao and Wang, Feng and Li, Zeming and Sun, Jian},
  journal={arXiv preprint arXiv:2107.08430},
  year={2021}
}

@inproceedings{fastrcnn,
  title={Fast r-cnn},
  author={Girshick, Ross},
  booktitle={Proceedings of the IEEE international conference on computer vision},
  pages={1440--1448},
  year={2015}
}

@article{hrnet,
  title={Deep high-resolution representation learning for visual recognition},
  author={Wang, Jingdong and Sun, Ke and Cheng, Tianheng and Jiang, Borui and Deng, Chaorui and Zhao, Yang and Liu, Dong and Mu, Yadong and Tan, Mingkui and Wang, Xinggang and others},
  journal={IEEE transactions on pattern analysis and machine intelligence},
  volume={43},
  number={10},
  pages={3349--3364},
  year={2020},
  publisher={IEEE}
}

@article{bigearth,
  title={Bigearthnet: A large-scale benchmark archive for remote sensing image understanding},
  author={Sumbul, Gencer and Charfuelan, Marcela and Demir, Beg{\"u}m and Markl, Volker},
  journal={arXiv preprint arXiv:1902.06148},
  year={2019}
}

@article{xview,
  title={xview: Objects in context in overhead imagery},
  author={Lam, Darius and Kuzma, Richard and McGee, Kevin and Dooley, Samuel and Laielli, Michael and Klaric, Matthew and Bulatov, Yaroslav and McCord, Brendan},
  journal={arXiv preprint arXiv:1802.07856},
  year={2018}
}

@inproceedings{iail,
  title={Can semantic labeling methods generalize to any city? the inria aerial image labeling benchmark},
  author={Maggiori, Emmanuel and Tarabalka, Yuliya and Charpiat, Guillaume and Alliez, Pierre},
  booktitle={2017 IEEE International geoscience and remote sensing symposium (IGARSS)},
  pages={3226--3229},
  year={2017},
  organization={IEEE}
}

@inproceedings{cowc,
  title={A large contextual dataset for classification, detection and counting of cars with deep learning},
  author={Mundhenk, T Nathan and Konjevod, Goran and Sakla, Wesam A and Boakye, Kofi},
  booktitle={European conference on computer vision},
  pages={785--800},
  year={2016},
  organization={Springer}
}

@article{FasterR-CNN,
  title={Faster R-CNN: Towards real-time object detection with region proposal networks},
  author={Ren, Shaoqing and He, Kaiming and Girshick, Ross and Sun, Jian},
  journal={IEEE transactions on pattern analysis and machine intelligence},
  volume={39},
  number={6},
  pages={1137--1149},
  year={2016},
  publisher={IEEE}
}

@inproceedings{Cascader-cnn,
  title={Cascade r-cnn: Delving into high quality object detection},
  author={Cai, Zhaowei and Vasconcelos, Nuno},
  booktitle={Proceedings of the IEEE conference on computer vision and pattern recognition},
  pages={6154--6162},
  year={2018}
}

@inproceedings{yolo,
  title={You only look once: Unified, real-time object detection},
  author={Redmon, Joseph and Divvala, Santosh and Girshick, Ross and Farhadi, Ali},
  booktitle={Proceedings of the IEEE conference on computer vision and pattern recognition},
  pages={779--788},
  year={2016}
}

@inproceedings{ssd,
  title={Ssd: Single shot multibox detector},
  author={Liu, Wei and Anguelov, Dragomir and Erhan, Dumitru and Szegedy, Christian and Reed, Scott and Fu, Cheng-Yang and Berg, Alexander C},
  booktitle={European conference on computer vision},
  pages={21--37},
  year={2016},
  organization={Springer}
}

@inproceedings{retinanet,
  title={Focal loss for dense object detection},
  author={Lin, Tsung-Yi and Goyal, Priya and Girshick, Ross and He, Kaiming and Doll{\'a}r, Piotr},
  booktitle={Proceedings of the IEEE international conference on computer vision},
  pages={2980--2988},
  year={2017}
}

@inproceedings{fcos,
  title={Fcos: Fully convolutional one-stage object detection},
  author={Tian, Zhi and Shen, Chunhua and Chen, Hao and He, Tong},
  booktitle={Proceedings of the IEEE/CVF international conference on computer vision},
  pages={9627--9636},
  year={2019}
}

@article{centernet,
  title={Objects as points},
  author={Zhou, Xingyi and Wang, Dequan and Kr{\"a}henb{\"u}hl, Philipp},
  journal={arXiv preprint arXiv:1904.07850},
  year={2019}
}

@inproceedings{detr,
  title={End-to-end object detection with transformers},
  author={Carion, Nicolas and Massa, Francisco and Synnaeve, Gabriel and Usunier, Nicolas and Kirillov, Alexander and Zagoruyko, Sergey},
  booktitle={European conference on computer vision},
  pages={213--229},
  year={2020},
  organization={Springer}
}

@article{deformable_detr,
  title={Deformable detr: Deformable transformers for end-to-end object detection},
  author={Zhu, Xizhou and Su, Weijie and Lu, Lewei and Li, Bin and Wang, Xiaogang and Dai, Jifeng},
  journal={arXiv preprint arXiv:2010.04159},
  year={2020}
}

@article{dino,
  title={Dino: Detr with improved denoising anchor boxes for end-to-end object detection},
  author={Zhang, Hao and Li, Feng and Liu, Shilong and Zhang, Lei and Su, Hang and Zhu, Jun and Ni, Lionel M and Shum, Heung-Yeung},
  journal={arXiv preprint arXiv:2203.03605},
  year={2022}
}

@inproceedings{rtdetr,
  title={Detrs beat yolos on real-time object detection},
  author={Zhao, Yian and Lv, Wenyu and Xu, Shangliang and Wei, Jinman and Wang, Guanzhong and Dang, Qingqing and Liu, Yi and Chen, Jie},
  booktitle={Proceedings of the IEEE/CVF conference on computer vision and pattern recognition},
  pages={16965--16974},
  year={2024}
}

@inproceedings{deim,
  title={Deim: Detr with improved matching for fast convergence},
  author={Huang, Shihua and Lu, Zhichao and Cun, Xiaodong and Yu, Yongjun and Zhou, Xiao and Shen, Xi},
  booktitle={Proceedings of the computer vision and pattern recognition conference},
  pages={15162--15171},
  year={2025}
}

@article{chen2019mmdetection,
  title={MMDetection: Open mmlab detection toolbox and benchmark},
  author={Chen, Kai and Wang, Jiaqi and Pang, Jiangmiao and Cao, Yuhang and Xiong, Yu and Li, Xiaoxiao and Sun, Shuyang and Feng, Wansen and Liu, Ziwei and Xu, Jiarui and others},
  journal={arXiv preprint arXiv:1906.07155},
  year={2019}
}

@article{mmengine2022,
  title   = {{MMEngine}: OpenMMLab Foundational Library for Training Deep Learning Models},
  author  = {MMEngine Contributors},
  howpublished = {\url{https://github.com/open-mmlab/mmengine}},
  year={2022}
}

@article{adcos,
  title={A benchmark dataset for aircraft detection in optical remote sensing imagery},
  author={Hu, Jianming and Zhi, Xiyang and Zhang, Bingxian and Shi, Tianjun and Cui, Qi and Sun, Xiaogang},
  journal={Remote Sensing},
  volume={16},
  number={24},
  pages={4699},
  year={2024},
  publisher={MDPI}
}

@article{hrplane,
  title={A benchmark dataset for deep learning-based airplane detection: HRPlanes},
  author={Bak{\i}rman, Tolga and Sertel, Elif},
  journal={International Journal of Engineering and Geosciences},
  volume={8},
  number={3},
  pages={212--223},
  year={2023},
  publisher={Murat YAKAR}
}

@inproceedings{skyfusion,
  title={Performance evolution of yolo models in remote sensing images},
  author={Hassan, Irfan and Xinyou, Zhang},
  booktitle={2024 21st International Computer Conference on Wavelet Active Media Technology and Information Processing (ICCWAMTIP)},
  pages={1--4},
  year={2024},
  organization={IEEE}
}

@INPROCEEDINGS{ucasaod,
  author={Zhu, Haigang and Chen, Xiaogang and Dai, Weiqun and Fu, Kun and Ye, Qixiang and Jiao, Jianbin},
  booktitle={2015 IEEE International Conference on Image Processing (ICIP)}, 
  title={Orientation robust object detection in aerial images using deep convolutional neural network}, 
  year={2015},
  volume={},
  number={},
  pages={3735-3739},
  doi={10.1109/ICIP.2015.7351502}}

@article{corsadd,
  title={Complex optical remote-sensing aircraft detection dataset and benchmark},
  author={Shi, Tianjun and Gong, Jinnan and Jiang, Shikai and Zhi, Xiyang and Bao, Guangzhen and Sun, Yu and Zhang, Wei},
  journal={IEEE Transactions on Geoscience and Remote Sensing},
  volume={61},
  pages={1--9},
  year={2023},
  doi={10.1109/TGRS.2023.3283137}
}

@inproceedings{carpk,
  title={Drone-based object counting by spatially regularized regional proposal network},
  author={Hsieh, Meng-Ru and Lin, Yen-Liang and Hsu, Winston H},
  booktitle={Proceedings of the IEEE international conference on computer vision},
  pages={4145--4153},
  year={2017}
}

@article{NWPU,
  title={Multi-class geospatial object detection and geographic image classification based on collection of part detectors},
  author={Cheng, Gong and Han, Junwei and Zhou, Peicheng and Guo, Lei},
  journal={ISPRS Journal of Photogrammetry and Remote Sensing},
  volume={98},
  pages={119--132},
  year={2014},
  publisher={Elsevier}
}

@article{vhrv,
  title={VHRV: Very High-Resolution Benchmark Dataset for Vessel Detection},
  author={B{\"u}y{\"u}kkanber, Furkan and Yanalak, Mustafa and Musao{\u{g}}lu, Nebiye},
  journal={Remote Sensing Applications: Society and Environment},
  volume={39},
  pages={101641},
  year={2025},
  publisher={Elsevier}
}

\end{document}